\documentclass[lettersize,journal]{IEEEtran}
\usepackage{amsmath,graphicx,hyperref}

\usepackage{cite}
\usepackage{amsmath,amssymb,amsfonts}
\usepackage{graphicx}
\usepackage{textcomp}
\usepackage{xcolor}
\usepackage{symbolDef}
\usepackage{subcaption}
\usepackage{theorem}
\def\BibTeX{{\rm B\kern-.05em{\sc i\kern-.025em b}\kern-.08em
    T\kern-.1667em\lower.7ex\hbox{E}\kern-.125emX}}

\newtheorem{theorem}{Theorem}
\newtheorem{assumption}{Assumption}
\newtheorem{proposition}{Proposition}
\newtheorem{lemma}{Lemma}
\newtheorem{definition}{Definition}

\def \QED {$\blacksquare$}
\usepackage{hyperref}
\usepackage{booktabs}
\usepackage{multirow}
\usepackage{algorithm}
\usepackage{algpseudocode}
\usepackage{pifont}
\newcommand{\mttTheta}[0]{\tilde{\mtTheta}} 
\newcommand{\first}[1]{\textbf{#1}}
\newcommand{\second}[1]{\textit{\underline{#1}}}

\usepackage{tikz}
\usetikzlibrary{shapes.geometric, arrows.meta, positioning, calc, backgrounds}

\algrenewcommand\algorithmicend{\textbf{end}}

\makeatletter
\algdef{SE}[FOR]{For}{EndFor}[1]{\algorithmicfor\ #1\ \algorithmicdo}{\algorithmicend}
\makeatother

\newenvironment{myproof}[1][$\!\!$]{{\noindent\bf Proof #1: }}
                         {\hfill\QED\medskip}

\input{mysymbols.sty}

\begin{document}
% Title.
% ------
\title{
Joint Precision Neural Networks: Task-Aware Dependency and Predictive Learning
}
% Single address.
% ---------------
\author{Andrea Cavallo~\IEEEmembership{Student Member,~IEEE,}  Samuel Rey~\IEEEmembership{Member,~IEEE,} Antonio G. Marques~\IEEEmembership{Senior Member,~IEEE,} and Elvin Isufi~\IEEEmembership{Senior Member,~IEEE}
\thanks{Andrea Cavallo and Elvin Isufi are with the INSY Department, Delft University of Technology. Samuel Rey and Antonio G. Marques are with the Department of Signal Theory and Communications, King Juan Carlos University. Emails:
    \{a.cavallo, e.isufi-1\}@tudelft.nl,
    \{samuel.rey.escudero, antonio.garcia.marques\}@urjc.es.
Work partially funded by the TU Delft AI Labs programme, NWO OTP GraSPA proposal \#19497, NWO VENI proposal 222.032, and by the  SURE-AI Centre grant \#357482, Research Council of Norway, the Spanish AEI (10.13039/501100011033) grants PID2022-136887NB-I00 \& PID2025-170000NB-I00, and the Community of Madrid via the Ellis Madrid Unit and grants URJC/CAM F1180 and TEC-2024/COM-89. Claude AI was used to assist in writing the manuscript and coding the simulations. The authors take full responsibility for the results of this paper.

}}

\maketitle
\begin{abstract}

Exploiting meaningful latent structures from data to solve downstream tasks is a fundamental challenge in signal processing and machine learning. While Principal Component Analysis (PCA) and coVariance Neural Networks (VNNs) successfully leverage the covariance matrix to process data, they inherently capture both direct and indirect correlations. The precision matrix (inverse covariance) overcomes this by explicitly encoding conditional independencies, making it largely studied in graphical lasso and graph topology identification.
However, finite-sample precision estimates are notoriously unstable, and regularized estimators remain task-agnostic.
In this work, our principal contribution is tackling the challenging problem of \emph{task-aware graph inference}. We propose Precision Neural Networks-Joint (PNN-Joint), a framework that jointly estimates a sparse, statistically grounded \emph{precision matrix} alongside \emph{graph neural network weights} via an alternating optimization scheme. As a foundational framework to support this, we introduce Precision Neural Networks (PNNs), a broader class of graph convolutional networks operating on precision estimators, and establish their spectral connections to PCA and VNNs alongside their stability to finite-sample errors. Extensive empirical evaluations on synthetic data, as well as real-world neuroimaging and motion sensor datasets, demonstrate that PNN-Joint yields highly interpretable task-aware graphs, exhibits remarkable robustness in low-data regimes, and consistently achieves the best or second-best performance among competitors on real-world tasks. 
\end{abstract}

\section{Introduction}

The analysis and processing of complex data to solve downstream tasks relies heavily on identifying and exploiting underlying statistical relationships.
For instance, modeling the co-activation patterns of brain areas is paramount for neurodegenerative disease detection~\cite{bashyam2020mri,yin2023anatomically}; financial time series forecasting largely depends on the correlated behaviors of different stocks~\cite{wang2022network,cardoso2020algorithms}; and sensor monitoring and processing requires understanding interlaced measurements~\cite{liao2022har,wang2023mhagnn}.

Traditionally, the sample covariance matrix has served as the cornerstone for understanding these structures. Techniques such as Principal Component Analysis (PCA)~\cite{Jolliffe2002pca} operate directly on the covariance eigenspace to perform dimensionality reduction and filter out noise. However, PCA suffers from several limitations: it acts as a rigid, task-agnostic linear operation, and its spectral decomposition is highly sensitive to finite-sample estimation errors, especially when
the gap between the eigenvalues is small or the data dimensionality is comparable to the sample size~\cite{deshp2016sparse,Jolliffe2002pca,Jolliffe2016PrincipalCA}.

To inject expressivity and stability into covariance-based processing, recent advancements in graph signal processing have introduced coVariance Neural Networks (VNNs)~\cite{sihag2022covariance,sihag2026learning}. VNNs treat the data features as nodes in a graph and the covariance matrix as the graph structure, and they apply graph convolutional filters and non-linearities to learn data-driven representations. Crucially, VNNs are trained end-to-end on a downstream task and provide improved stability to finite-sample errors. Despite this, processing data via the covariance matrix presents a structural drawback as the covariance captures both direct and indirect correlations. Consequently, the resulting graph is often dense due to spurious connections that do not reflect true or direct interactions.

A more principled proxy for latent structure is the precision matrix, the inverse of the covariance. Under multivariate Gaussian assumptions, the precision matrix explicitly encodes conditional independencies, forming a sparse Gaussian graphical model where edges represent direct dependencies~\cite{lauritzen1996graphical}. Unfortunately, leveraging the precision matrix as an inductive bias in neural architectures is challenging. The unregularized sample precision matrix is highly susceptible to noise and is not even defined if the number of samples is smaller than the sample dimension, since the sample covariance is singular.
The unregularized sample precision matrix is highly susceptible to noise and is uncomputable in high-dimensional regimes. Established estimators like the graphical lasso~\cite{friedman2007lasso} elegantly mitigate this by promoting structural sparsity, which filters out spurious noise to yield a statistically robust, interpretable and more parsimonious topology.
However, they operate entirely unsupervised. The resulting precision matrix, though statistically sound, is task-agnostic and may discard information crucial for downstream classification or regression tasks.

Simultaneously, the literature on joint graph and Graph Neural Network (GNN) learning has sought to estimate task-aware topologies end-to-end~\cite{franceschi2019learning, fatemi2021slaps}. However, these approaches predominantly rely on heuristic topological penalties, lacking formal convergence guarantees toward the true underlying graph. Furthermore, they are rarely grounded in the fundamental statistical moments of the observed data.

In this work, our principal contribution is bridging the gap between rigorous statistical estimation and task-aware deep learning by tackling the coupled optimization landscape of precision estimation. We propose Precision Neural Network-Joint (PNN-Joint), a framework that alternates between estimating a sparse, graphical lasso-regularized precision matrix and updating the network weights, ensuring the inferred topology is optimized for the downstream task. As a theoretical foundation to support this, we formalize Precision Neural Networks (PNNs), a broader class of convolutional GNNs that utilize the precision matrix
 as their underlying shift operator to directly exploit the conditional independencies of the data.
The main contributions of this paper are as follows.

\noindent \textbf{(C1) Joint Task-Aware Estimation.} We formulate a joint design and propose an alternating optimization algorithm that minimizes a downstream task loss and a graphical lasso penalty. This allows the model to learn a sparse, task-aware precision matrix anchored to the true statistics of the data.

\noindent \textbf{(C2) Precision Neural Networks.} We formally define the broader PNN framework and establish its theoretical connections to PCA and VNNs. We show that PNNs operate by learning a data-driven frequency response over the precision eigenvalues, generalizing the rigid spectral filtering of PCA.

\noindent \textbf{(C3) Theoretical Guarantees.} We derive stability bounds for PNNs under finite-sample estimation errors, showing that their output perturbation decreases with the inverse square root of the sample size, and we prove that the precision matrix estimated by our joint framework converges to the true one at a rate dependent on its sparsity, up to a bias term that quantifies the task-induced deviation from the graphical lasso solution.

\noindent \textbf{(C4) Empirical Validation.} We extensively evaluate PNNs on synthetic benchmarks and real-world datasets, encompassing neuroimaging and human activity recognition. Our results demonstrate that PNN-Joint consistently ranks among the top 2 models for prediction quality, exhibits strong stability in low-sample regimes, and yields interpretable task-aware graphs.

A preliminary version of this manuscript appeared in the short paper~\cite{cavallo2025pnns}. In this work, we extend the theoretical analysis to provide stability bounds for PNNs, and provide comprehensive experiments on synthetic and real-world datasets.
All proofs are collected in the Appendix.

\section{Preliminaries}
\label{sec:preliminaries}

We introduce the background information of the paper, covering PCA, VNNs and precision matrix estimation.

\subsection{Covariance Matrix and PCA}

Let $\mtX \in \reals^{N\times T}$ be a zero-mean data matrix whose columns $\vcx_i \in \reals^N$ are observations of a random variable $\vcx$ with covariance $\mtC_0 = \mathbb{E}[\vcx\vcx^\Tr]$.
Since $\mtC_0$ is generally unknown, it is approximated by the sample estimate $\mthC = \mtX\mtX^\Tr / T$, whose eigendecomposition is given by $\mthC = \mtV \mtLambda \mtV^\Tr$. Here, the diagonal matrix $\mtLambda \in \reals^{N \times N}$ collects the eigenvalues $\{ \lambda_i \}_{i=1}^N$ and $\mtV \in \reals^{N \times N}$ contains the orthogonal eigenvectors.
This covariance matrix and its sample estimate form the basis for several widely adopted data processing techniques.
Consider also prediction targets $\mtY \in \mathbb{R}^{N_y\times T}$, with $N_y$ the prediction size (e.g., $N_y=N$ for node-level classification or regression, $N_y=1$ for graph signal classification or regression, etc.).
Our goal is to predict $\mtY$ from $\mtX$ exploiting relational information from the covariance structure of the data.

One prominent covariance-based data processing technique is PCA, which applies a transformation $\mttX = \mtV^\Tr \mtX$ to diagonalize the sample covariance, yielding $\mttX \mttX^\Tr /T = \mtLambda$.
For dimensionality reduction, PCA selects the $\tilde N$ eigenvectors corresponding to the largest eigenvalues, mapping the data to $\mttX_{(\tilde N)} = [\mtV]_{:,1:\tilde N}^\Tr\mtX$.
 This operation acts as an ideal high-pass spectral filter on the covariance eigenvalues, where $h_\textnormal{PCA}(\lambda_i) = \mathbf{1}[\lambda_i \geq \lambda_{\tilde N}]$ (cf. Figure~\ref{fig:filters}).
However, PCA is prone to finite-sample instabilities, since the estimation error of the $i$-th principal component is inversely proportional to
the eigengap $|\lambda_{i} - \lambda_{i+1}|$, meaning close eigenvalues in low-data regimes ($N \approx T$) degrade the projection (see~\cite[Lemma 1]{cavallo2024sparse}).
If errors artificially inflate a smaller eigenvalue, PCA selects suboptimal components.
Additionally, PCA maximizes variance, which may discard structurally relevant information if the data contains high-variance noise, and remains agnostic to the downstream task, potentially worsening performance.

\subsection{(Sparse) Covariance Filters and Neural Networks}

\begin{figure}
    \centering
    \includegraphics[width=\linewidth]{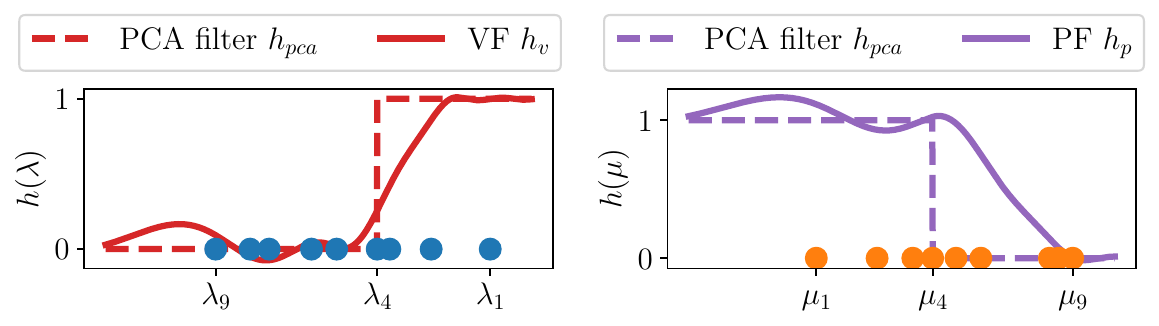}
    \caption{Spectral responses of PCA, covariance and precision filters. PCA applies a high-pass filter in the covariance spectrum, selecting only the eigenvectors corresponding to the largest eigenvalues, which translates to a low-pass filter in the precision spectrum.
    Covariance filters (VFs) and precision filters (PFs) learn smooth polynomial functions in the same spaces. The figure illustrates two example polynomials of order 10 approximating the PCA filtering.
    }
    \label{fig:filters}
\end{figure}

\begin{figure*}
    \centering
    \includegraphics[width=.8\linewidth]{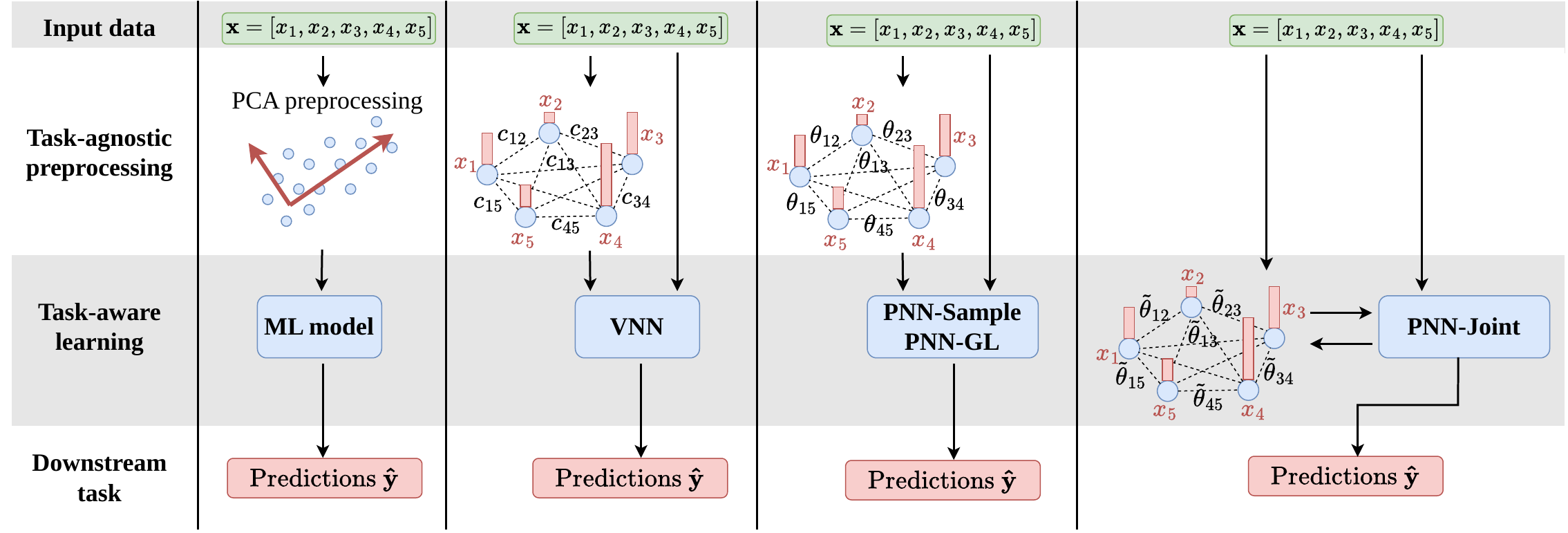}
    \caption{
    Comparison of covariance and precision-based methods for downstream tasks. Traditional pipelines preprocess the data via (task-agnostic) PCA and apply a separate ML model for the task. VNNs and PNNs perform task-aware covariance/precision processing, while PNN-Joint learns both the precision and the neural weights jointly in a task-aware manner.
    }
    \label{fig:method}
\end{figure*}

To address these limitations while maintaining expressive processing, covariance filters~\cite{sihag2022covariance} treat the data features as graph nodes and the covariance $\mthC$ as the underlying graph structure (cf. Figure~\ref{fig:method}). A covariance filter is a polynomial matrix function transforming input signal $\vcx$ into output $\vcu$ as 
\begin{equation}\label{eq:cov_filter}
    \vcu = \mtH(\mthC)\vcx = \sum_{k=0}^K h_k \mthC^k \vcx.
\end{equation}
This operation can be seen in the spectral domain, by multiplying by matrix $\mtV^\Tr$: $\vctu = \mtV^\Tr \vcu = \sum_{k=0}^K h_k \mtLambda^k \mtV^\Tr \vcx$.
Since $\mtLambda$ is diagonal, this corresponds to a polynomial frequency response for each eigenvalue $\lambda_i$: $h(\lambda_i) = \sum_{k=0}^K h_k \lambda_i^k$. This demonstrates the analogy with PCA, as both filter the covariance spectrum, and specific coefficients $h_k$ can perfectly recover PCA's high-pass behavior for a sufficiently large order $K$ (see~\cite[Theorem 1]{sihag2022covariance} and Figure~\ref{fig:filters}).

Covariance filters fundamentally improve stability to estimation errors. The filter's spectral Lipschitz constant $P$, defined as $|h(\lambda_i) - h(\lambda_j)| \leq P |\lambda_i - \lambda_j|$, limits variations in the frequency response. By bounding $P$, the output discrepancy caused by sample estimation errors scales proportionally with $\mathcal{O}(1/\sqrt{T})$. This entirely removes PCA's restrictive dependency on the eigengap.
To further reduce spurious correlations and computational overhead, sparse covariance filters~\cite{cavallo2024sparse} replace the dense matrix $\mthC$ with a sparsified estimate, improving both efficiency and robustness to spurious correlations, while maintaining finite-sample stability guarantees.

While covariance filters are robust, they are restricted to linear transformations. (Sparse) VNNs incorporate multi-layer architectures
 and non-linearities to increase expressive power~\cite{sihag2022covariance,cavallo2024sparse}.
Grouping the $F_\ell$ feature channels at layer $\ell$ into a signal matrix $\mtX^{(\ell)} \in \reals^{N \times F_\ell}$, the layer update rule is:
\begin{equation}\label{eq:vnn}
    \mtX^{(\ell)} = \sigma\left( \sum_{k=0}^K \mthC^k \mtX^{(\ell-1)} \mtW_k^{(\ell)} \right),
\end{equation}
where $\mtX^{(\ell-1)} \in \reals^{N \times F_{\ell-1}}$ is the input from the previous layer (with $\mtX^{(0)} = \vcx$), $\mtW_k^{(\ell)} \in \reals^{F_{\ell-1} \times F_\ell}$ is the learnable weight matrix for the $k$-th filter order, and $\sigma(\cdot)$ is an element-wise non-linear activation function.
The network parameters are optimized end-to-end to minimize a task-specific loss. While VNNs excel at covariance-based learning, the covariance matrix may not be the optimal proxy for latent structures, motivating the use of the precision matrix.

\subsection{Precision Matrix and Graphical Lasso}

The precision matrix is the inverse of the true covariance, $\mtTheta_0 = \mtC_0^{-1}$, with sample estimate $\hat \mtTheta= \mthC^{-1}$ and eigendecomposition $\hat \mtTheta = \mtV \mtM \mtV^\Tr$ where $\mtV$ are the covariance eigenvectors and $\mtM = \mtLambda^{-1}$ contains the inverse covariance eigenvalues $\{ \mu_i = \lambda_i^{-1} \}_{i=1}^N$.
Unlike the covariance matrix, which contains both direct and indirect dependencies, the precision matrix explicitly encodes
 conditional dependencies for Gaussian data.
An entry $[\mtTheta_0]_{ij} = 0$ indicates that variables $[\vcx]_i$ and $[\vcx]_j$ are conditionally independent given all other variables. This property is highly advantageous for topology inference~\cite{giannakis2018topoid} and causal discovery~\cite{vowels2022d}, among others.

Because the finite-sample estimate $\hat\mtTheta$ is often noisy and requires many samples for proper estimation, especially in high-dimensional regimes, the graphical lasso (GL) is used to infer a \emph{sparse} precision matrix under multivariate Gaussian assumptions~\cite{friedman2007lasso}. Specifically, it solves the following $\ell_1$-regularized maximum likelihood problem:
\begin{equation}\label{eq:gl}
    \mtTheta_\text{GL} = \arg\min_{\mtTheta \succ 0}  -\log \det(\mtTheta) + \text{tr}(\mthC\mtTheta) + \beta_\text{GL} \|\mtTheta_{\bar{\mathcal{D}}}\|_1 ,
\end{equation}
where $\mtTheta \succ 0$ enforces symmetric positive definiteness, $\|\mtTheta_{\bar{\mathcal{D}}}\|_1$ is the element-wise $\ell_1$-norm of the off-diagonal entries, and $\beta_\text{GL} > 0$ controls the structural sparsity.
Increasing $\beta_\text{GL}$ drives more off-diagonal elements to exactly zero, isolating the strongest conditional dependencies. Additionally, this regularization acts as a structural shrinkage operator, ensuring the estimate remains invertible and well-conditioned even in high-dimensional regimes ($N > T$) where the standard sample covariance is rank-deficient~\cite{banerjee2008model}.

\section{Precision Neural Networks (PNNs)}
\label{sec:pnns}

We introduce the PNN framework and analyze the connections with PCA and VNNs.
The joint estimation of a task-aware precision matrix and the PNN weights is addressed in Section~\ref{sec:optimization}.
We compare PNNs, PCA and VNNs in Figure~\ref{fig:method}.

\subsection{Model Definition}

PNNs are convolutional GNNs operating on the precision matrix. To formalize PNNs,
 we first introduce the concept of a precision filter and then extend it to a multi-layer neural network architecture.

\smallskip
\noindent \textbf{Precision Filters (PF).}
A precision filter follows the covariance filter in~\eqref{eq:cov_filter}, replacing the covariance with a precision estimate $\mtTheta$: $\vcu = \mtH(\mtTheta)\vcx = \sum_{k=0}^K h_{k} \mtTheta^k \vcx$.
Since the support of $\mtTheta$
encodes the conditional independencies among the entries of $\bbx$, this filter is a graph convolution over the conditional-dependence graph of the data.
This yields two structural advantages. First, $\mtTheta$ often admits a sparse structure, enabling efficient sparse matrix-vector multiplications. Second, the filter explicitly leverages conditional independencies, propagating information only between directly dependent variables.

\smallskip
\noindent \textbf{PNN Architecture.}
To enhance the expressive capability of a single linear precision filter, we define PNNs following the VNN principle in~\eqref{eq:cov_filter}, where we replace $\mthC$ with $\mtTheta$:
\begin{equation}\label{eq:pnn}
    \mtX^{(\ell)} = \sigma\left( \sum_{k=0}^K \mtTheta^k \mtX^{(\ell-1)} \mtW_k^{(\ell)} \right).
\end{equation}
The final representation $\mtX^{(L)}$ is then either aggregated or passed to a readout layer to produce the prediction for the downstream task. We denote the resulting end-to-end map by $\Phi(\mtX, \mtTheta, \vcw)$, where $\vcw$ collects all trainable parameters $\{\mtW_k^{(\ell)}\}_{k,\ell}$.
Given a training set $(\mtX_\textnormal{tr}, \mtY_\textnormal{tr})$, these parameters are learned by minimizing
\begin{equation}\label{eq:training_pnn}
    \min_\vcw \mathcal{L}_\textnormal{task}\left(\mtY_\textnormal{tr}, \Phi(\mtX_\textnormal{tr},  \mtTheta, \vcw)\right) + \beta\|\vcw\|_2^2,
\end{equation}
where $\mathcal{L}_\textnormal{task}$ is a task-specific loss, such as cross-entropy for classification or mean squared error for regression, and $\beta \geq 0$ controls the strength of the squared $\ell_2$ regularization on the trainable weights.

\subsection{Spectral Analysis and Connection to PCA and VNNs}

For population matrices and sample estimators, the precision matrix $\mtTheta = \mtC^{-1}$ shares its eigenbasis $\mtV$ with the covariance matrix, but has inverse eigenvalues, i.e., $\mu_i=\lambda_i^{-1}$.
Following the graph signal processing paradigm~\cite{Gama_2020}, we define the \textit{precision Fourier transform} that projects a signal $\vcx$ onto this basis: $\vctx = \mtV^\Tr \vcx$.
Different precision estimators might obtain different eigenvectors from those of the sample covariance. In the following, we consider the settings in which they are the same (i.e., population and sample estimators).

\smallskip \noindent \textbf{Connection to PCA.}
The precision Fourier transform is mathematically equivalent to the principal component representation used by PCA.
The precision filter applied to the $i$-th frequency component simplifies to
\begin{equation}\label{eq:freq_resp}
    [\vctu]_i = \sum_{k=0}^K h_k \mu_i^k [\vctx]_i = h(\mu_i) [\vctx]_i.
\end{equation}
This relation demonstrates that PNNs operate by modulating the PCA projections of the data. However, unlike standard PCA, which applies a rigid low-pass filter in the precision spectrum, a precision filter learns a flexible, data-driven polynomial frequency response $h(\mu)$ evaluated on the precision eigenvalues $\mu_i$. A PF can approximate the PCA transform while also learning more expressive task-driven functions.

\smallskip
\noindent \textbf{Connection to VNNs.}
Because PNNs and VNNs share the identical Fourier basis $\mtV$, they operate in the same spectral space, but with reversed eigenvalue ordering (cf. Figure~\ref{fig:filters}). Since the spectrum contains $N$ eigenvalues, any covariance filter response can be exactly reproduced by a precision filter of order $K = N-1$ (Lagrange interpolation over $\{\mu_i\}_{i=1}^N$), and vice-versa. Hence, for sufficiently large filter orders, PNNs and VNNs can represent the same set of spectral transformations, whereas for small $K$ the two parametrizations span different function classes.
Nevertheless, in practice, the covariance and precision matrices incorporate different inductive biases which might lead to different downstream performance.

\section{Optimization Algorithm}
\label{sec:optimization}

PNNs require an estimate of the precision matrix. A simple approach is to compute $\mtTheta$ by inverting the sample covariance $\mthC$ (\textbf{PNN-Sample}), but this is known to yield poor estimates, particularly in high-dimensional or low-sample regimes~\cite{friedman2007lasso}. Alternatively, one may employ regularized approaches such as GL (cf.~\eqref{eq:gl}), which directly estimate $\mtTheta$ under structural assumptions like sparsity (\textbf{PNN-GL}). Although these approaches typically provide more reliable precision matrices, they remain task-agnostic, since the estimation does not incorporate the downstream objective. As a result, the obtained precision matrix may still be suboptimal for the learning problem.

% Joint optimization formulation
To address this limitation, we propose a task-aware framework where we \emph{jointly estimate the PNN weights and the precision matrix} (\textbf{PNN-Joint}).
Given a training set $(\mtX_\textnormal{tr}, \mtY_\textnormal{tr})$ and a convolutional GNN architecture $\Phi$, we propose solving
\begin{align}
    \min_{\vcw,\mtTheta} & \;\; \alpha \mathcal{L}_\textnormal{task}(\mtY_\textnormal{tr}, \Phi(\mtX_\textnormal{tr},  \mtTheta, \vcw)) + (1 - \alpha) \ccalL_\textnormal{GL} (\mtX_\textnormal{tr},  \mtTheta) \nonumber \\ \textnormal{s.t.} & \quad \mtTheta \mathrel{\succ} 0, \quad  \|\mtTheta\|_2 \leq M, \label{eq:initial_prob}
\end{align}
where $\mathcal{L}_\textnormal{task}$ is the task loss from \eqref{eq:training_pnn}, $\ccalL_\textnormal{GL}$ is the GL objective in~\eqref{eq:gl}, and $\|\cdot\|_2$ denotes the spectral norm such that $\|\mtTheta\|_2 \leq M$ ensures bounded eigenvalues.
The parameter $\alpha \in [0,1]$ balances the influence of task-driven and model-driven terms, yielding a precision matrix $\mtTheta$ that supports the PNN in the downstream task.
The loss in~\eqref{eq:initial_prob} is non-convex and multi-objective, making its optimization complex.  
While a naive gradient descent over $(\mtTheta, \vcw)$ is possible (we call this approach \textbf{PNN-Naive}), it poses a coupled and challenging landscape, leading to complex and potentially unstable gradients~\cite{jin2020graph}. This justifies our relaxation and alternating optimization.

\subsection{Relaxed Formulation and Optimization}

To partially alleviate these issues, we start by adopting the following relaxed formulation
\begin{align}\label{eq:joint_prob}
    \min_{\vcw,\mtTheta,\mttTheta}  &\quad \alpha \mathcal{L}_\textnormal{task}(\mtY_\textnormal{tr}, \Phi(\mtX_\textnormal{tr},  \mttTheta, \vcw))
    + (1 - \alpha) \ccalL_\textnormal{GL} (\mtX_\textnormal{tr},  \mtTheta) \nonumber \\
    &+ \frac{\gamma}{2} \| \mtTheta - \mttTheta \|_F^2 \quad \textnormal{s.t.}  \quad \mtTheta \mathrel{\succ} 0, \quad \| \mtTheta\|_2 \leq M,
\end{align}
where the auxiliary variable $\mttTheta$ decouples $\mathcal{L}_\textnormal{task}$ from $\mathcal{L}_\textnormal{GL}$ through a quadratic penalty.
In the limit when $\gamma \to \infty$, the constraint $\mtTheta = \mttTheta$ is enforced exactly, and problem~\eqref{eq:joint_prob} reduces to~\eqref{eq:initial_prob}.
This alternative formulation is amenable to alternating optimization~\cite{jin2020graph,tenorio2023robust}.
We approach it by sequentially solving the following steps for $i=1, \ldots, I_{\text{epochs}}$ iterations.

\vspace{2mm}
\noindent
\textbf{Step 1.}
We update $\mtTheta$ with $\mttTheta^{(i)}$ fixed by solving
\begin{align}
    \mtTheta^{(i+1)} = &\argmin_{\mtTheta} (1 - \alpha) \ccalL_\textnormal{GL} (\mtX_\textnormal{tr},  \mtTheta) + \frac{\gamma^{(i)}}{2} \| \mtTheta - \mttTheta^{(i)} \|_F^2 \nonumber \\
    &\textnormal{s.t.} \quad \mtTheta \mathrel{\succ} 0, \quad \| \mtTheta\|_2 \leq M. \label{eq:step1}
\end{align}
This step amounts to a strictly convex optimization problem closely related to GL, which can be solved efficiently with existing algorithms.
We adopt a proximal gradient descent approach~\cite{beck2017first,navarro2024fair} where gradient descent is applied to the smooth terms in \eqref{eq:step1}, followed by the proximal operator of the $\ell_1$ norm (soft-thresholding of the off-diagonal elements), and finally projection onto the cone of positive semidefinite matrices and norm clipping.
While this update is only an approximation of the exact proximal step (the projections are not composable with soft-thresholding), such heuristics are common in practice~\cite{navarro2024fair} and we observed no convergence issues in our experiments. The analysis in Section~\ref{sec:theory} refers to the exact minimizer of \eqref{eq:step1}.

\vspace{2mm}
\noindent
\textbf{Step 2.}
Next, we update $\mttTheta$ while fixing $\vcw^{(i)}$ and $\mtTheta^{(i+1)}$.
This requires solving
\begin{align}\label{eq:step2}
    \mttTheta^{(i+1)} = \argmin_{\mttTheta} \;& \alpha  \mathcal{L}_\textnormal{task}(\mtY_\textnormal{tr}, \Phi(\mtX_\textnormal{tr}, \mttTheta, \vcw^{(i)})) \nonumber \\
    &+ \frac{\gamma^{(i)}}{2} \| \mtTheta^{(i+1)} - \mttTheta  \|_F^2.
\end{align}
In practice, this step is implemented via gradient descent, where the gradient is a linear combination of $(\mtTheta^{(i+1)}-\tilde{\mtTheta})$ and $\nabla_{\tilde{\mtTheta}}\mathcal{L}_\textnormal{task}$. 
Since $\nabla_{\tilde{\mtTheta}}\mathcal{L}_\textnormal{task}$ is not symmetric in general, after each block of updates of $\tilde{\mtTheta}$ we project it onto the set of symmetric matrices via $\tilde{\mtTheta} \leftarrow (\tilde{\mtTheta} + \tilde{\mtTheta}^\Tr)/2$.

\vspace{2mm}
\noindent
\textbf{Step 3.}
Finally, we update the PNN weights while keeping the previous estimates $\mtTheta^{(i+1)}$ and $\mttTheta^{(i+1)}$ fixed.
This amounts to
\begin{equation}\label{eq:step3}
    \vcw^{(i+1)} = \argmin_{\vcw} \alpha \mathcal{L}_\textnormal{task}(\mtY_\textnormal{tr}, \Phi(\mtX_\textnormal{tr},  \mttTheta^{(i+1)}, \vcw)),
\end{equation}
which corresponds to standard NN training and can be performed with, e.g., the Adam optimizer~\cite{kingma2014adam}.

The overall procedure is summarized in Algorithm~\ref{alg:alternating}, where we let $\mathcal{L}(\vcw,\mtTheta,\tilde{\mtTheta})$ denote the objective function in~\eqref{eq:joint_prob}. In Algorithm~\ref{alg:alternating}, $\eta>0$ is the step size and $\epsilon>0$ is a small constant that keeps the matrix inverse in the gradient of $-\log\det(\cdot)$ well defined.
Compared to the naive (global) implementation, PNN-Naive, PNN-Joint has the following merits:
i) the alternating optimization finds an approximate solution to the original problem through simpler problems;
ii) the step to estimate the precision is decoupled from the PNN training, leading to a biased GL problem where the first term is the standard GL objective, while the quadratic term acts as a ridge-type penalty that shrinks the solution toward $\mttTheta^{(i)}$, with $\gamma^{(i)}$ controlling the strength of the bias. This is a convex problem that guarantees the existence of the global minimizer, and enables assessing the quality of the estimated precision, which we will explore in the next section;
iii) this also allows for accommodating discrepancies between the precision matrix used in the task and the assumption that data follows a Gaussian Markov random field (GMRF), since the two matrices have to be close but do not need to be the same.
On the negative side, the joint formulation introduces the hyperparameter $\gamma$ which requires careful tuning. Following standard practice in penalty methods, we let the hyperparameter $\gamma^{(i)}$ depend on the iteration $i$
to allow for a lower regularization at the beginning (small $i$) for better exploration, and larger regularization towards the end (large $i$) to ensure that $\mtTheta, \mttTheta$ are close enough. 
Finally, the joint optimization scheme incurs longer training time due to the alternating updates and the additional variable.

\begin{algorithm}[t]
\caption{Iterative optimization of $\mtTheta$, $\tilde{\mtTheta}$, and $\vcw$}
\label{alg:alternating}
\begin{algorithmic}[1]
\footnotesize
    \State Initialize $\mtTheta^{(1)}$, $\tilde{\mtTheta}^{(1)}$, $\vcw^{(1)}$
    \For{$i = 1, \dots, I_{\text{epochs}}$}
        \State $\mtTheta^{(i+1)} \gets \mtTheta^{(i)}$,\ $\tilde{\mtTheta}^{(i+1)} \gets \tilde{\mtTheta}^{(i)}$,\ $\vcw^{(i+1)} \gets \vcw^{(i)}$
        \For{$j = 1, \dots, I_{\Theta}$}
            \State $\mtTheta^{(i+1)} \gets \mtTheta^{(i+1)} - \eta \big((1-\alpha)(\mthC -(\mtTheta^{(i+1)}+\epsilon \mtI)^{-1})$
            \Statex \hspace{7.5em} $+\, \gamma^{(i)} (\mtTheta^{(i+1)} - \tilde{\mtTheta}^{(i)})\big)$
            \State $\mtTheta^{(i+1)} \gets \text{soft\_threshold}(\mtTheta^{(i+1)}, \eta(1-\alpha)\beta_\text{GL})$
            \State $\mtTheta^{(i+1)} \gets \text{psd\_projection}(\mtTheta^{(i+1)})$
            \State $\mtTheta^{(i+1)} \gets M/\max(M, \|\mtTheta^{(i+1)}\|_2) \mtTheta^{(i+1)}$
        \EndFor
        \For{$k = 1, \dots, I_{\tilde{\Theta}}$}
            \State $\tilde{\mtTheta}^{(i+1)} \gets \tilde{\mtTheta}^{(i+1)} - \eta \nabla_{\tilde{\mtTheta}} \mathcal{L}(\vcw^{(i)},\mtTheta^{(i+1)},\tilde{\mtTheta}^{(i+1)})$
        \EndFor
        \For{$l = 1, \dots, I_{w}$}
            \State $\vcw^{(i+1)} \gets \text{adam\_update}(\vcw^{(i+1)}, \mathcal{L}(\vcw^{(i+1)},\mtTheta^{(i+1)},\tilde{\mtTheta}^{(i+1)}))$
        \EndFor
    \EndFor
    \State Return $\mtTheta^{(I_\textnormal{epochs}+1)}$, $\tilde{\mtTheta}^{(I_\textnormal{epochs}+1)}$, $\vcw^{(I_\textnormal{epochs}+1)}$
\end{algorithmic}
\end{algorithm}

\section{Theoretical Analysis}
\label{sec:theory}

PNNs rely on an estimate of the precision matrix, which inevitably deviates from the true precision and may therefore introduce perturbations that affect downstream performance.
We first analyze how finite-sample errors propagate through the output of a PF when the precision is obtained either by inverting the sample covariance or via GL.
We then study the joint setting, bounding the error incurred when the precision matrix is estimated together with the filter weights, before extending the analysis to full PNN architectures.
Our analysis relies on the notion of Lipschitz PFs, introduced next.
\begin{definition}
\label{def:lipschitz}
    A PF is Lipschitz with constant $P$ on an interval $\mathcal{I} \subset \reals$ if, for every pair $\mu, \mu' \in \mathcal{I}$, its frequency response satisfies: $|h(\mu)-h(\mu')| \leq  P|\mu - \mu'|$. In the sequel, $\mathcal{I}$ is any bounded interval containing the eigenvalues of the true precision $\mtTheta_0$ and of the precision estimate under consideration; since $h$ is a polynomial, such a $P$ always exists.
\end{definition}
The Lipschitz constant $P$ is informative of the rate of variability of the filter's frequency response. A larger $P$ allows sharper variations, making the filter more expressive in discriminating close eigenvalues, but also more sensitive to eigenvalue perturbations.
We will also use the following assumption.
\begin{assumption}\label{as:data_distr}
    The observed data follows a Gaussian distribution $\vcx \sim \mathcal{N}(\mathbf{0}, \mtC_0)$, where $\mtC_0$ has bounded eigenvalues in the range $[\lambda_\textnormal{min},\lambda_\textnormal{max}]$ with $\lambda_\textnormal{min}>0, \lambda_\textnormal{max}<\infty$ and $\mtTheta_0 = \mtC_0^{-1}$ has $S$ non-zero off-diagonal entries.
\end{assumption}

We provide the stability bounds next, using the notation $\mathcal{O}_P$, which describes the asymptotic behavior of a sequence of random variables, indicating that the sequence is bounded with probability approaching 1 as $N,T$ approach infinity.
We begin by providing stability results for precision filters, and we later show their extension to the full PNN architecture.

\subsection{Stability of Task-Agnostic PFs}

We start with the sample precision estimator, which provides the simplest baseline for constructing a PF.

\begin{proposition}\label{th:stability_sample}
    Let $\hat{\mtTheta} = \mthC^{-1}$ be the sample precision estimate from $T \geq N$ samples and $\mtH(\cdot)$ be a precision filter with Lipschitz constant $P$ on an interval containing the eigenvalues of $\hat{\mtTheta}$ and $\mtTheta_0$.
    Assume $T$ is large enough such that $\|\mthC - \mtC_0\|_2 \leq \lambda_\textnormal{min}/2$.
    Under Assumption~\ref{as:data_distr}, the output perturbation of the precision filter is bounded as
    \begin{equation}\nonumber
        \| \mtH(\hat{\mtTheta})\vcx - \mtH(\mtTheta_0)\vcx \|_2 = \mathcal{O}_P \left( \frac{PN}{\sqrt{T}} \|\vcx\|_2 \right).
    \end{equation}
\end{proposition}

Proposition~\ref{th:stability_sample} shows that, as the graph grows larger, the perturbation bound increases at a rate proportional to $N$, leading to less stable filters.
 In contrast, stability improves with the number of samples as $1/\sqrt{T}$, since more observations
 yield a more accurate precision estimate. Nevertheless, the scaling in Proposition~\ref{th:stability_sample} highlights the challenging nature of high-dimensional regimes.

Then, we proceed to analyze the stability of a PF operating on the graphical lasso-based precision $\mtTheta_\textnormal{GL}$ from~\eqref{eq:gl}.
\begin{proposition}\label{th:stability_gl}
    Let $\mtTheta_{\textnormal{GL}}$ be the graphical lasso precision estimate from $T$ samples and $\mtH(\cdot)$ be a precision filter with Lipschitz constant $P$ on an interval containing the eigenvalues of $\mtTheta_{\textnormal{GL}}$ and $\mtTheta_0$. Fix $\beta_\textnormal{GL} \asymp \sqrt{\log N / T}$.
    Under Assumption~\ref{as:data_distr}, the output error of the precision filter under finite-sample estimation error is bounded as
    \begin{equation}\nonumber
        \| \mtH(\mtTheta_{\textnormal{GL}})\vcx - \mtH(\mtTheta_0)\vcx \|_2 = \mathcal{O}_P\Big(\tfrac{P}{\sqrt{T}}\sqrt{(N+S)\log{N}}\,\|\vcx\|_2 \Big) .
    \end{equation}
\end{proposition}

Compared to Proposition~\ref{th:stability_sample}, Proposition~\ref{th:stability_gl} highlights a more convenient dependency w.r.t. the data dimension $N$. The improvement stems from the sharper estimation of $\mtTheta_{\textnormal{GL}}$. When the true precision is sparse ($S =\mathcal{O}( N)$), the estimation error scales approximately as $\| \mtTheta_{\textnormal{GL}} - \mtTheta_0 \|_F = \mathcal{O}_P\big(\sqrt{N\log N / T}\big)$, instead of the $N/\sqrt{T}$ scaling associated with the sample precision estimate. Thus, finite-sample perturbations grow more mildly with $N$. For dense precision matrices ($S\approx N^2$), this sparsity advantage vanishes and the bound recovers, up to a logarithmic factor, the order of Proposition~\ref{th:stability_sample}.

\subsection{Stability of PF-Joint}

Next, we study the stability of the PFs and PNN, where the precision matrix $\mttTheta$ is estimated while accounting for the downstream task through Algorithm~\ref{alg:alternating}.
To this end, we extend the result in our conference precursor~\cite[Theorem 1]{cavallo2025pnns} to bound the distance from the minimizer of~\eqref{eq:step1} at iteration $i$, i.e., $\mtTheta^{(i)}$, and the true precision $\mtTheta_0$, under the assumption that the matrices $\mtTheta^{(i)}$ and $\tilde{\mtTheta}^{(i)}$ are close enough, a condition that can be controlled through the hyperparameter $\gamma$.

\begin{theorem}\label{th:precision_convergence_ext}
    Let $\alpha \in [0,1)$ and let $\mtTheta^{(i)}$ denote the minimizer of the objective in~\eqref{eq:step1} with $\beta_\textnormal{GL} = Q\sqrt{\log N/T}$, for a sufficiently large constant $Q$, at iteration $i$, let $M \geq 1/\lambda_\textnormal{min}$, and let $\| \mtTheta^{(i)} - \tilde{\mtTheta}^{(i)} \|_F \leq \xi$.
    Under Assumption~\ref{as:data_distr}, it holds that
    \begin{equation}\nonumber
        \| \mtTheta^{(i)} - \mtTheta_0 \|_F = \mathcal{O}_P \left( \sqrt{\frac{(N+S)\log N}{T}} + \frac{\gamma}{1-\alpha}\,\xi \right) .
    \end{equation}
\end{theorem}

Theorem~\ref{th:precision_convergence_ext} shows that, by using the additional assumption on the iterate difference $\| \mtTheta^{(i)} - \tilde{\mtTheta}^{(i)} \|_F \leq \xi$ (at the iteration $i$ under consideration), the effect of finite-sample estimation errors on the estimated precision decreases with rate $\mathcal{O}(T^{-1/2})$, while the overall bound is still bounded away from zero by the term $\frac{\gamma}{1-\alpha}\xi$, which quantifies the task-induced deviation of $\mtTheta^{(i)}$ from the graphical lasso solution.

\begin{theorem}\label{th:stability}
    Let $\mtTheta^{(i)},\mttTheta^{(i)}$ be as in Theorem~\ref{th:precision_convergence_ext}, with $\mttTheta^{(i)}$ symmetric, and $\mtH(\cdot)$ be a precision filter with Lipschitz constant $P$ on an interval containing the eigenvalues of $\mttTheta^{(i)}$ and $\mtTheta_0$.
    The perturbation of the precision filter under finite-sample estimation error is bounded as
    \begin{align}\nonumber
        \| \mtH(\tilde{\mtTheta}^{(i)})\vcx &- \mtH(\mtTheta_0)\vcx \|_2\\
        &=  \mathcal{O}_P \Big( \Big[ P\sqrt{\tfrac{(N+S)\log N}{T}} + P\big(1+\tfrac{\gamma}{1-\alpha}\big)\xi \Big] \|\vcx\|_2 \Big) .\nonumber
    \end{align}
\end{theorem}

Theorem~\ref{th:stability} shows that the stability of a PF-Joint depends on the finite-sample estimation errors with a rate analogous to the standard graphical lasso, but is additionally affected by the distance $\xi$ which measures the discrepancy between the task-aware precision estimate and the graphical lasso-like estimate. If the task requires a matrix that is close to the true precision, then $\xi \ll 1$ and the bound is not significantly affected. Moreover, $\xi$ can be controlled by varying $\gamma$ in the optimization algorithm, keeping in mind that the bound depends on $\gamma\xi$.

\subsection{Stability of PNNs}

With the filter bounds in place, we analyze the stability of a complete PNN architecture.

\begin{proposition}
\label{th:pnn_stability}
    Consider a PNN $\Phi(\vcx, \mtTheta, \vcw)$ of $L$ layers with $F_\ell = F$ for $\ell=1,\ldots,L$, precision filters with bounded operator norm $\|\mtH(\cdot)\|_2 \leq 1$ (i.e., $|h(\mu)| \leq 1$ on the relevant interval), and nonlinearities $\sigma(\cdot)$ such that $|\sigma(a)-\sigma(b)|\leq |a-b|$ and $\sigma(0)=0$.
    Let $\bar{\mtTheta}$ be the precision estimate used by the PNN (either $\hat \mtTheta$, $\mtTheta_\textnormal{GL}$ or $\mttTheta^{(i)}$) and let $\Delta$ be the largest bound of all precision filters in the PNN under the assumptions of Propositions~\ref{th:stability_sample}, \ref{th:stability_gl} or Theorem \ref{th:stability}. For a generic data sample $\vcx$, it holds
    \begin{align}
 \|\Phi(\vcx, \bar{\mtTheta}, \vcw) - \Phi(\vcx, \mtTheta_0, \vcw)\|_2 = \nonumber \mathcal{O}_P \left( LF^{L-1}\Delta \|\vcx\|_2 \right).
    \end{align}
\end{proposition}

\begin{myproof}
The proof follows by replacing the PF bound in Theorems~\ref{th:stability_sample}, \ref{th:stability_gl} or~\ref{th:stability} into the bound in \cite[Theorem 4]{gama2020stability}.
\end{myproof}

Proposition~\ref{th:pnn_stability} shows that the filter instabilities are amplified by the number of layers $L$ in the PNN and by the feature size $F$ at each layer, reflecting the expressivity-stability tradeoff common in GNNs~\cite{gama2020stability}: increasing $L$ and $F$ generally improves expressivity, but reduces stability due to the larger number of filters in the NNs, each affected by instability.

\begin{figure*}[t]
    \begin{subfigure}{\textwidth}
        \centering
        \includegraphics[width=.8\linewidth]{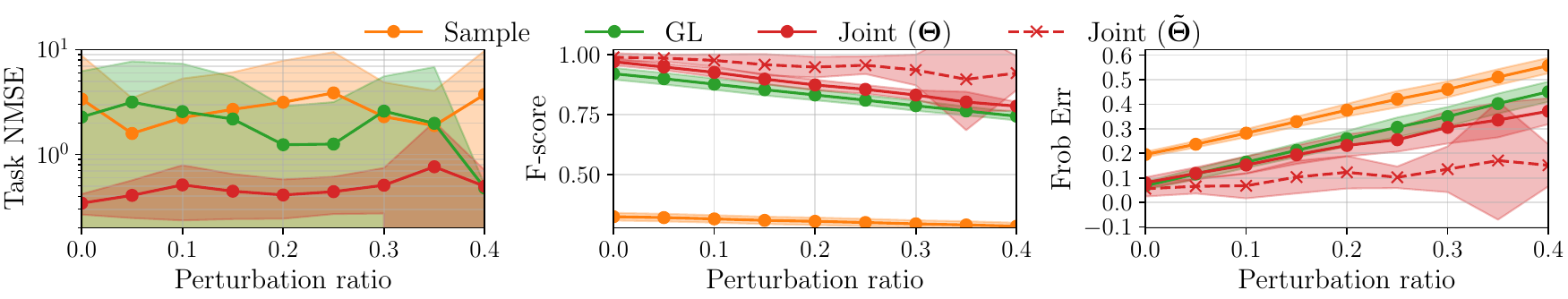}
        \caption*{}
    \end{subfigure}
    \begin{subfigure}{\textwidth}
        \centering
        \includegraphics[width=.8\linewidth]{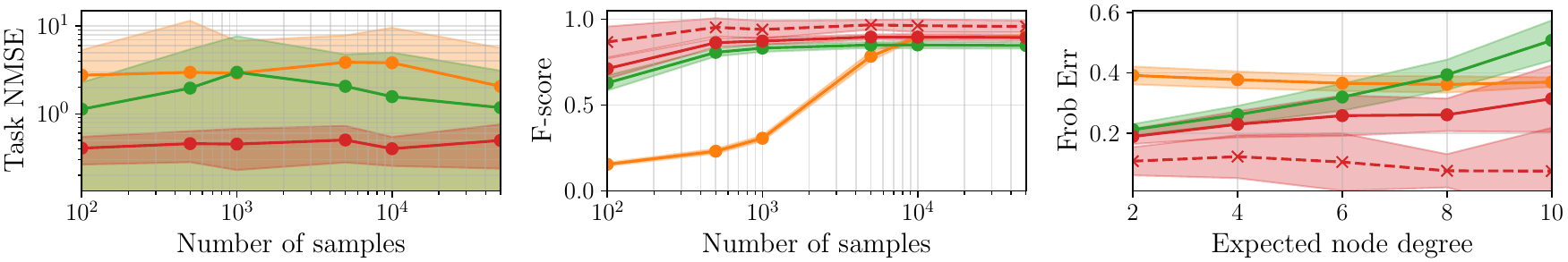}
        \caption*{}
    \end{subfigure}
    \caption{Regression task on synthetic datasets under three settings: (i-top row) increasing structural mismatch between the statistical precision matrix $\mtTheta_0$ and the task precision matrix $\mtTheta_\textnormal{task}$; (ii-bottom row, left and center panel) increasing number of samples; (iii-bottom right) increasing sparsity of the true precision matrix. To assess downstream performance, we report the normalized mean squared error (NMSE). To evaluate precision recovery, we report the scale-invariant normalized Frobenius error and the F1-score of the recovered support for both $\mtTheta$ and $\mttTheta$. }
    \label{fig:synth_regr_A}
\end{figure*}

\section{Numerical Results}
\label{sec:experiments}

We evaluate PNNs numerically with four objectives.
\textbf{(O1)} On synthetic data, verify that PNN-Joint recovers the precision matrices while achieving competitive task performance; this validates the central claim that structure learning and task learning need not be traded off.
\textbf{(O2)} On real datasets, compare PNN-Joint against competitive baselines, to establish that the gains in (O1) carry over to settings where the generative model is unknown.
\textbf{(O3)} Measure how task performance degrades with finite-sample errors, to corroborate the finite-sample stability result of Section~\ref{sec:theory}.
\textbf{(O4)} Inspect the learned task-aware precision matrices to explain how the task shapes the estimated graph.
We make our code available on GitHub\footnote{\url{https://github.com/andrea-cavallo-98/PNN}}, together with the full details of the simulation setup.

\subsection{Synthetic Data: Regression Task \textbf{(O1)}}

We generate a synthetic regression task with known precision matrix $\mtTheta_0 \in \reals^{N \times N}$. We sample $T$ observations, stacked in the data matrix $\mtX \in \reals^{N \times T}$, whose columns are i.i.d.\ $\mathcal{N}(\mathbf{0}, \mtTheta_0^{-1})$.
To introduce a controlled structural mismatch, we generate a task-specific precision matrix $\mtTheta_\textnormal{task}$ by randomly perturbing $\mtTheta_0$, rewiring a fraction of its edges, which we refer to as the perturbation ratio.
The regression targets $\mtY \in \reals^{N \times T}$ are then generated by applying a polynomial precision filter of order $K=2$ over the perturbed graph $\mtTheta_\textnormal{task}$. For each input signal $\vcx \in \reals^N$ (i.e., a column of $\mtX$), the target $\vcy \in \reals^N$ is $\vcy = \sum_{k=0}^K h_k \mtTheta_\textnormal{task}^k \vcx + \mathbf{n}$,
where $\{h_k\}_{k=0}^K$ are the filter coefficients, and $\mathbf{n} \sim \mathcal{N}(\mathbf{0}, \sigma_n^2 \mathbf{I})$ is white Gaussian noise. The filter coefficients are drawn independently (i.i.d.) from a uniform distribution $\mathcal{U}(0,1)$ and normalized such that $\sum_{k=0}^K h_k = 1$.
 Finally, the noise variance $\sigma_n^2$ is scaled to satisfy a predetermined
Signal-to-Noise Ratio (SNR) in dB.

\smallskip \noindent \textbf{Experimental Setup.}
Unless otherwise stated, we generate the synthetic data considering $N=50$ variables, $T=1000$ samples, an expected node degree of $4$, a perturbation ratio of $0.2$ and an SNR of 10 dB.
We compare PNN-Sample, PNN-GL and PNN-Joint using the same architecture in all synthetic settings. The PNN has two graph-filter layers with feature size $F=16$, filter order $K=1$, ReLU activation and dropout $0.5$. We initialize $\mtTheta^{(1)}$ and $\tilde{\mtTheta}^{(1)}$ to identity. We train the neural weights with Adam using learning rate $0.01$ and weight decay $10^{-4}$ for at most $1000$ epochs, with early stopping patience $30$. For PNN-GL, we set $\beta_{\textnormal{GL}} = 0.4\sqrt{\log N/T}$ and use $\eta=0.001$ for the precision updates. For PNN-Joint, we use the same $\beta_{\textnormal{GL}}$, $\eta=0.005$, $\alpha=0.75$ and schedule the quadratic penalty $\gamma$ from $\gamma_{\min}=0.001$ to $\gamma_{\max}=0.5$. Each epoch (i.e., outer iteration) performs $10$ inner iterations. We repeat each experiment over $20$ random realizations.

Figure~\ref{fig:synth_regr_A} reports the mean performance and standard deviations across realizations. We evaluate downstream performance through the normalized mean squared error (NMSE), and precision recovery through the F1-score of the recovered support (after thresholding entries with magnitude below 0.1) and the scale-invariant normalized Frobenius error for both $\mtTheta$ and $\mttTheta$. We consider three synthetic scenarios: i) increasing the link perturbation ratio between $\mtTheta_0$ and $\mtTheta_\textnormal{task}$; ii) increasing the number of samples; and iii) varying the graph sparsity through the mean node degree.

\smallskip \noindent \textbf{Discussion.}
PNN-Joint consistently achieves the lowest downstream NMSE across all perturbation levels and sample sizes, demonstrating that task-aware joint optimization directly translates to improved task performance.
Beyond the downstream task, this task-awareness also enhances the recovery of the underlying graph structure. Across various sample sizes, perturbation ratios and expected node degrees, PNN-Joint yields better Frobenius errors and F-scores for both $\mtTheta$ and $\mttTheta$ compared to the baselines.
Interestingly, the auxiliary matrix $\mttTheta$ generally shows better recovery metrics than the primary precision matrix $\mtTheta$. This suggests that $\mttTheta$ effectively captures and leverages task-specific structural signals, while $\mtTheta$ acts as an anchor to maintain consistency with data statistics.

In contrast, the baselines struggle without this task-aware feedback loop. PNN-Sample produces the highest downstream regression errors and performs poorly at precision recovery unless provided with a massive number of samples. Meanwhile, PNN-GL yields more stable recovery across different sample sizes due to its sparsity regularization. However, because it is entirely task-agnostic, it ultimately settles for suboptimal results in both downstream task performance and precision recovery.
Moreover, both PNN-Sample and PNN-GL produce large standard deviations across runs, indicating that initialization is crucial for their proper functioning, and their task-agnostic precision estimation is sometimes detrimental for the downstream task. PNN-Joint, on the other hand, performs more consistently across runs, confirming the advantage of task-aware precision estimation.

\begin{table*}[t]
\centering
\caption{Regression error (MAE), classification accuracy and percentage of non-zero elements in the estimated precision matrix (NZ, after thresholding entries with magnitude lower than 0.1) for real-data experiments. \first{First} and \second{second} best-performing methods are highlighted. For PNN-Joint, NZ is computed on $\tilde \mtTheta$.}
\resizebox{\textwidth}{!}{
\begin{tabular}{c|cc|cc|cc|cc|cc|cc}
\toprule
\multirow{2}{*}{\textbf{Method}} & \multicolumn{2}{c|}{\textbf{ADNI1}} & \multicolumn{2}{c|}{\textbf{ADNI2}} & \multicolumn{2}{c|}{\textbf{PPMI}} & \multicolumn{2}{c|}{\textbf{ABIDE}} & \multicolumn{2}{c|}{\textbf{MHEALTH}} & \multicolumn{2}{c}{\textbf{REALDISP}} \\
 & MAE & NZ (\%)  & MAE & NZ (\%) & MAE & NZ (\%) & MAE & NZ (\%) & Acc (\%) & NZ (\%) & Acc (\%) & NZ (\%) \\
\midrule
MLP & 9.39$\pm$1.21 & -  & 7.99$\pm$0.35 & -  & 7.26$\pm$0.41 & -  & 3.88$\pm$0.38 & -  & 82.1$\pm$1.1 & -  & 75.0$\pm$4.7 & -  \\
PCA+MLP & 5.59$\pm$0.47 & -  & \second{5.40$\pm$0.54} & -  & 6.66$\pm$0.49 & -  & 3.73$\pm$0.18 & -  & 81.7$\pm$0.7 & -  & 76.4$\pm$0.7 & -   \\
SPCA+MLP & \second{5.23$\pm$0.30} & -  & 5.50$\pm$0.42 & -  & \second{6.57$\pm$0.56} & -  & \second{3.62$\pm$0.28} & -  & 81.4$\pm$3.4 & -  & 72.0$\pm$1.4 & -   \\
VNN~\cite{sihag2022covariance}   & 5.31$\pm$0.42 & - & 5.62$\pm$0.29 & - & 6.90$\pm$0.48 & - & 5.89$\pm$0.79 & - & 87.3$\pm$2.8 & - & 79.3$\pm$3.1 & -  \\
S-VNN~\cite{cavallo2024sparse}   & 5.35$\pm$0.39 & - & 5.78$\pm$0.12 & - & 7.64$\pm$0.59 & - & 6.46$\pm$1.52 & - & \first{88.1$\pm$2.4} & - & 80.7$\pm$0.9 & -  \\
LDS~\cite{franceschi2019learning}    & 5.35$\pm$0.46 & - & 5.57$\pm$0.27 & - & 6.88$\pm$0.48 & - & 5.70$\pm$0.58 & - & 76.2$\pm$1.3 & - & 73.5$\pm$1.7 & -  \\
\midrule
PNN-Sample   & 5.37$\pm$0.39 & 64$\pm$1 & 5.54$\pm$0.34 & 52$\pm$1 & 6.97$\pm$0.45 & 60$\pm$1 & 5.23$\pm$0.62 & 80$\pm$1 & 86.1$\pm$2.1 & 56$\pm$0 & 72.1$\pm$1.4 & 56$\pm$0  \\
PNN-GL      & 5.29$\pm$0.46 & 18$\pm$1 & 5.66$\pm$0.32 & 17$\pm$0 & 6.69$\pm$0.52 & 13$\pm$0 & 4.69$\pm$0.36 & 31$\pm$0 & 85.0$\pm$2.5 & 50$\pm$0 & 79.2$\pm$3.2 & 26$\pm$0  \\
PNN-Naive & 5.82$\pm$0.65 & 95$\pm$5 & 6.13$\pm$0.77 & 96$\pm$3 & 6.87$\pm$0.68 & 99$\pm$1 & 4.02$\pm$0.36 & 99$\pm$1 & 86.1$\pm$2.4 & 56$\pm$0 & \second{81.7$\pm$2.5} & 26$\pm$0 \\
PNN-Joint    & \first{5.00$\pm$0.31} & 74$\pm$24 & \first{5.33$\pm$0.28} & 94$\pm$2 & \first{6.44$\pm$0.53} & 75$\pm$23 & \first{3.39$\pm$0.27} & 49$\pm$5 & \second{88.0$\pm$1.6} & 56$\pm$0 & \first{83.7$\pm$1.3} & 26$\pm$0  \\
\bottomrule
\end{tabular}
}
\label{tab:real_data}
\end{table*}

\begin{figure*}[t]
    \centering
    \includegraphics[width=\linewidth]{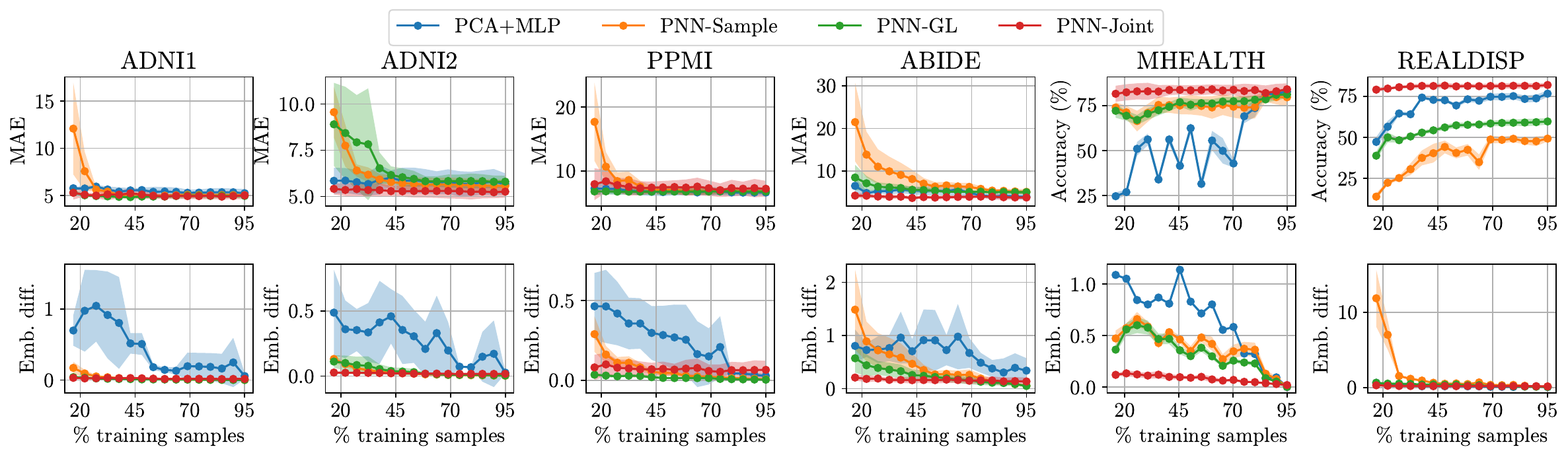}
    \caption{Results on real datasets under finite-sample perturbations. The first row shows the task performance, the second the normalized Frobenius norm of the difference of the generated embeddings.
    }
    \label{fig:stability}
\end{figure*}

\begin{figure}[t]
    \centering
    \includegraphics[width=\linewidth]{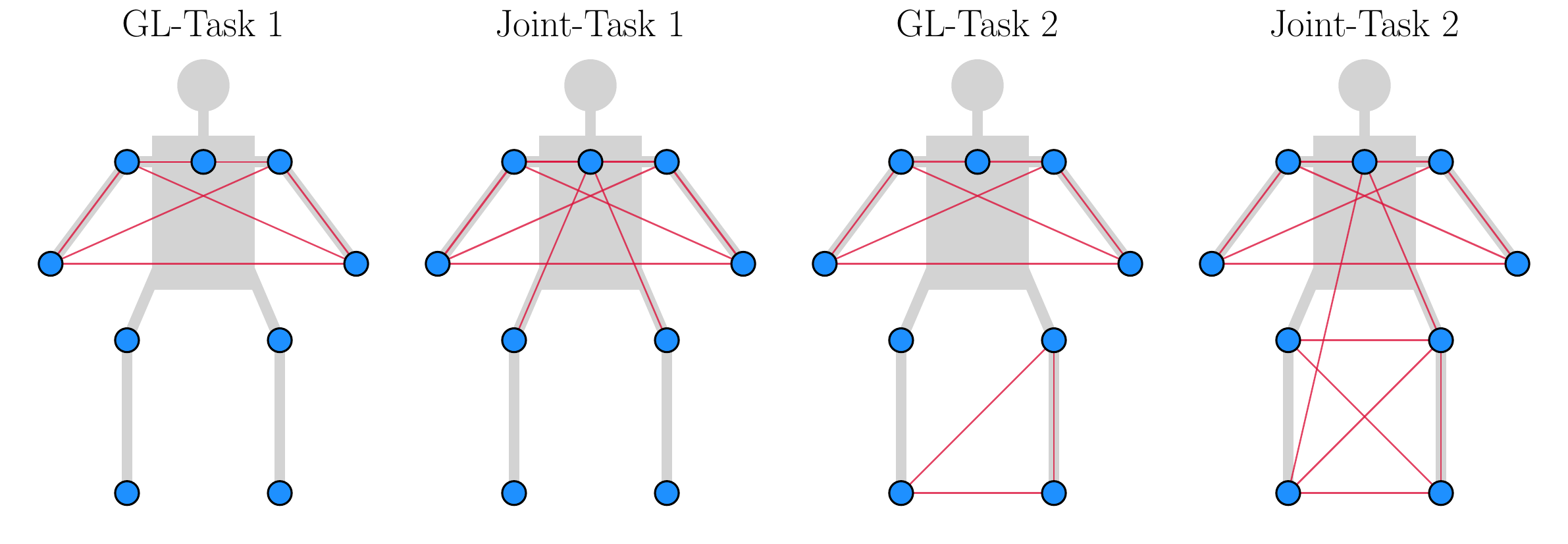}
    \caption{Visualization of the graphs learned on different subsets of REALDISP. Task 1 involves activities mainly concerned with the upper body, while Task 2 considers actions that more clearly depend on leg movements. The learned graphs reflect this distinction: on Task 1, PNN-GL only learns edges between upper-body sensors and PNN-Joint adds one edge from the chest to the upper legs, while on Task 2 more edges across sensors on the legs appear.}
    \label{fig:realdisp_graphs}
\end{figure}

\subsection{Real Data \textbf{(O2)}}

To assess the performance of the proposed PNNs on real data, we consider six datasets spanning two application domains: brain imaging and wearable motion sensing.

\smallskip
\noindent \textbf{Brain Datasets.}
We focus on the task of predicting a patient's age from cortical thickness measurements extracted from MRI scans (regression task). We utilize three medical imaging datasets containing patients suffering from specific conditions. The \textbf{ADNI} dataset~\cite{jack2008alzheimer} targets Alzheimer's disease. We use two subsets differing in their collection protocols, ADNI1 and ADNI2, comprising 801 and 1142 patients respectively, with measurements across 68 brain regions extracted via FreeSurfer~\cite{fischl2012freesurfer}. The \textbf{PPMI} dataset~\cite{marek2011parkinson} focuses on Parkinson's disease, providing data for 1704 patients evaluated over the same 68 cortical regions. Finally, the \textbf{ABIDE} dataset~\cite{craddock2013neuro} addresses autism spectrum disorder, containing clinical information for 1035 patients with features extracted across 62 brain regions following the Mindboggle protocol~\cite{klein2017mindboggling}.

\smallskip
\noindent \textbf{Motion Sensor Datasets.} For human activity recognition, we consider the task of classifying the action a person is performing from wearable sensor measurements (classification task). We use two datasets containing 3-axis accelerometer measurements. The \textbf{MHEALTH} dataset~\cite{banos2014mhealth} includes data from 10 subjects performing 12 distinct actions, yielding 6 accelerometer measurements from devices placed on the right wrist and left ankle. The \textbf{REALDISP} dataset~\cite{misc_realdisp_activity_recognition_dataset_305} encompasses 17 subjects and provides 27 accelerometer measurements collected across 9 body locations, from which we classify the top 10 most frequently performed actions. For both datasets, we follow the preprocessing of~\cite{tello2023good}, segmenting the continuous sensor data into temporal windows of size 128 (2.56 seconds) with a 50\% overlap, and we maintain strictly disjoint sets of subjects across the training, validation, and test.

\smallskip \noindent \textbf{Experimental Setup.}
As baselines, we consider: i) an MLP trained directly on the raw features; ii) PCA- and sparse PCA-based pipelines, where the data are first projected and then fed to an MLP for prediction; iii) covariance-based neural architectures, namely VNN~\cite{sihag2022covariance} and S-VNN~\cite{cavallo2024sparse} with soft thresholding; and iv) LDS, a joint graph and GNN learning method~\cite{franceschi2019learning}. We compare these methods against four PNN variants: PNN-Sample, PNN-GL, PNN-Naive, and PNN-Joint.
For MLP, PCA and SPCA baselines, we use the following hyperparameter grids: number of layers $L\in \{1,2,3\}$, feature size $F\in \{16,32\}$ and the number of latent components $D \in \{5,10,20\}$ for brain datasets and $D \in \{2,4,6\}$ for motion datasets.
For VNNs and S-VNNs, we use the following hyperparameter grids: layers $L \in \{1,2,3\}$, feature size $F=16$, order $K \in \{1,2\}$, threshold for S-VNN $\tau = \beta_0 \sqrt{\log N / T}$ where $\beta_0 \in \{ 0.01,0.1,1 \}$.
For PNNs, we use the following hyperparameter grids: layers $L \in \{1,2,3\}$, feature size $F=16$, order $K \in \{1,2\}$, $\eta = 0.005$ for PNN-Joint and $\eta=0.001$ for PNN-GL, $\beta_{\text{GL}} = \beta_0 \sqrt{\log N / T}$ where $\beta_0 \in \{ 0.01,0.1,1 \}$, $\gamma \in \{1, 10\}$, $\alpha=0.5$.
We train all models for 1000 epochs (i.e., outer iterations), we use Adam optimizer with learning rate 0.01 and weight decay $10^{-5}$ and we use batch normalization. For each outer iteration, we perform 10 inner iterations.
We repeat all experiments 5 times and report mean results and standard deviations.

\smallskip \noindent \textbf{Discussion.}
As shown in Table~\ref{tab:real_data}, PNN-Joint delivers the most consistent performance across all datasets. This confirms that guiding the joint estimation of the precision matrix and neural network weights with task-specific feedback is highly effective. In contrast, baseline methods that lack this integration fall short. For instance, VNNs and S-VNNs perform comparably to PNN-GL on most datasets, as all three rely on purely task-agnostic estimates of the data's statistical moments. PNN-Sample performs even worse on most datasets, as it struggles to filter out the spurious noise inherent in unregularized sample precision matrices. Furthermore, PNN-Joint generally outperforms LDS, demonstrating that explicitly modeling precision information provides an advantage over joint estimation methods that ignore it entirely. Finally, the fact that simple PCA preprocessing beats an MLP trained on raw features suggests that relevant predictive information is organized along the covariance eigenvectors, which supports methods that process them directly such as VNNs and PNNs.
Compared to PNN-Naive, which also includes task-awareness in the precision estimation, but is not optimized in an alternating fashion, PNN-Joint produces a better downstream performance on all datasets, and estimates generally sparser precision matrices. This corroborates that the alternating optimization strategy allows more control on the graph structure, since PNN-Naive struggles to prune edges, while at the same time guaranteeing a more consistent accuracy on the task. Nevertheless, the increase in performance comes at the cost of additional computational complexity, since PNN-Joint has more variables to optimize.

Regarding the structural properties of the learned graphs, PNN-Joint generally produces sparse precision matrices, though they remain dense on certain datasets. This density occurs because, in our main experiments, we optimized the hyperparameters strictly to maximize downstream task performance rather than to enforce sparsity. However, the framework is highly flexible. If recovering a sparse graph is a priority, the hyperparameters can be explicitly tuned to balance sparsity with task accuracy. For example, using an alternative parameter configuration on the ADNI2 dataset yields a significantly sparser estimate (33\% $\pm$ 37\% non-zero entries) with only a minor increase in error (MAE of 5.36 $\pm$ 0.28). This demonstrates a favorable trade-off, allowing practitioners to achieve much sparser representations with minimal cost to predictive performance.
The number of non-zero entries varies considerably across runs, since each run uses a different data split (and therefore a different sample covariance), so the learned graph adapts to the samples available; the task performance, however, remains stable across runs, which suggests that graphs with different sparsity levels can support the downstream task equally well.

\subsection{Stability \textbf{(O3)}}
We assess the empirical impact of the stability to finite-sample estimations discussed in Section~\ref{sec:theory}.
On the 6 real-world datasets, we train PNNs on the full training set, and, during test, we re-compute precision estimates on subsets of the training set with fewer samples. For PNN-GL, this implies re-running the GL algorithm, while for PNN-Joint, this requires running steps 1 and 2 in Algorithm~\ref{alg:alternating}, while keeping the neural network weights fixed. This introduces finite-sample errors in the estimations of the precision matrices and allows analyzing the performance of PNNs with fixed weights on these perturbed variants.
We use the best hyperparameter configuration from Table~\ref{tab:real_data} for each method and dataset.

Figure~\ref{fig:stability} shows that PNN-GL and PNN-Joint achieve remarkable stability to finite-sample estimation errors, providing consistent performance with reduced sample size and small embedding perturbations. PNN-Sample is stable for large enough sample sizes, but suffers more in low-data regimes as the inverse operation amplifies spurious noise. Finally, PCA generally performs the worst, since it has no mechanism to counteract finite-sample errors that might significantly affect principal component estimation.
The improved stability of PNN-GL w.r.t. PNN-Sample aligns with the improved bound in Proposition~\ref{th:stability_gl} compared to Proposition~\ref{th:stability_sample}, confirming the advantage of GL-based precision estimation in low-sample regimes.
Compared to PNN-GL, the results show that PNN-Joint achieves similar or better stability. This is in apparent contrast with the stability bounds, which suggest that PNN-Joint should be less stable.
However, the bounds are worst-case scenarios where the task can bias the estimated precision away from the true precision. In practice, we observe a comparable performance.
Finally, all PNNs are more stable than PCA, thus retaining the finite-sample stability that motivated VNNs.

\subsection{Interpretability \textbf{(O4)}}

PNN-Joint can be interpreted in two ways: i) learning a precision matrix that helps the downstream task, and ii) using the task as prior information towards a better graph estimation.
Consequently, observing the learned graph offers valuable interpretability into the model's decision-making process, as well as understanding of what graph structure is meaningful for the data at hand.
To investigate this, we partitioned the REALDISP dataset into two behaviorally distinct subsets. Task 1 consists of upper-body actions that require minimal leg movement: rowing, trunk twist (arms outstretched) and trunk twist (elbows bent). Conversely, Task 2 comprises actions where lower-body movement is more relevant: walking, jogging and running.

To visualize the learned structures, we aggregate the node-level connections into broader body-location edges. Because each physical sensor outputs three distinct measurements (represented as three separate nodes in the graph), we define the overall edge weight between any two physical body locations as the average of the edge weights connecting their corresponding sensor nodes. We consider $\tilde \mtTheta$ and apply a final hard threshold of 0.2.

As illustrated in Figure~\ref{fig:realdisp_graphs}, PNN-GL estimates no edges involving leg sensors in Task 1, and PNN-Joint only connects the chest sensor to one upper-leg sensor, as leg movements are not relevant towards discriminating the actions performed.
On Task 2, instead, both models estimate edges across leg sensors, indicating that these are more informative in this setting. Moreover, PNN-Joint estimates more edges than PNN-GL, and edges between leg and upper body sensors, suggesting that a denser graph structure is more helpful for the task.

\section{Related Work}
\label{sec:related_works}

The joint estimation of graph structures and GNN parameters has been previously explored, typically by adding to a task-driven loss a tailored regularizer or parametrization of the graph. Regarding how the graph is modeled, \cite{franceschi2019learning} parametrizes it via independent Bernoulli distributions, \cite{chen2019deep} derives the topology from attention-driven node similarities, \cite{wang2021graph} fits a Stochastic Block Model, and \cite{sahbi2021learning} optimizes the adjacency matrix directly under spectral regularizers. Regarding how the graph and the weights are optimized, \cite{franceschi2019learning} casts the problem as a bilevel program solved with generic hypergradient methods, whereas \cite{chen2019deep,wang2021graph} alternate between updating the graph and the GNN weights. In all cases, the graph-learning subproblem is treated as a smooth black box, so neither its structure nor non-smooth sparsity-promoting penalties such as the $\ell_1$ norm are exploited, which is the gap we address here.
Alternatively, end-to-end training frameworks estimate the graph concurrently with the task by leveraging auxiliary self-supervised feature reconstruction losses~\cite{fatemi2021slaps} or metapath-based similarities for heterogeneous graphs~\cite{zhao2021heterogeneous}.
Some works jointly estimate a topology and a graph filter, rather than a complete GNN, for system identification~\cite{natali2020topology,egilmez2018graph}, while another line of works proposes a joint GNN and graph learning scheme for robustness to topology perturbations~\cite{rey2023robust,tenorio2023robust} or adversarial attacks by estimating a corrected graph~\cite{zhang2022asgnn,jin2020graph}.
Despite their empirical success, these existing methods share fundamental theoretical limitations. Notably, they rely on heuristic graph penalties and most of them lack guarantees relating the learned graph to the statistical structure of the data.
Furthermore, these graph estimation processes are not grounded in the statistical moments or covariance structures of the observed data. 

A recent line of work injects task information into the estimator. \cite{zhang2026learning} regularizes a data-fitting term with the solution of a lower-level task (a network game or dynamical equilibrium) in a bilevel program, while \cite{saboksayr2021online,batreddy2023robust,raghuvanshi2025task} use class labels to learn discriminative graphs. These methods target recovery of a graph rather than prediction, and their task is not a learned model. PNN-Joint sits between these and the joint GNN/graph-learning methods above, since the task is a trainable GNN, yet the estimate remains anchored to the GMRF likelihood, which is what enables the convergence result of Theorem~\ref{th:precision_convergence_ext}.

The advantages of GNNs to process covariance information were first shown by VNNs in~\cite{sihag2022covariance}, which analyzed their stability to finite-sample perturbations and connections to PCA. These advantages, along with their transferability across resolutions~\cite{sihag2023transferablility}, led to successful applications of VNNs to interpretable brain age gap prediction~\cite{sihag2024explainable,sihag2025disentangling}, brain computer interface data processing~\cite{roy2026covariance}, time series forecasting~\cite{cavallo2024stvnn} and label-scarce scenarios~\cite{cavallo2025covariancescattering}.

Nevertheless, VNNs and their extensions operate on regularized estimates of the covariance matrix that do not directly account for the downstream labels.

\section{Conclusion}

In this paper, we formulated PNN-Joint, an alternating optimization scheme to integrate downstream task objectives with precision matrix estimation via a graphical lasso penalty. 
To contextualize our formulation, we introduced PNNs, i.e., GNNs relying on the precision matrix as a graph shift operator, and framed PNN-Joint as a particular case to address the inherent limitations of sample precision estimates and the task-agnostic nature of standard regularizers. 
Our theoretical analysis established that PNNs share the spectral foundations of PCA and VNNs, and offer mathematically proven stability bounds against finite-sample estimation errors. Our evaluation of PNN-Joint on synthetic scenarios, medical neuroimaging, and motion sensor datasets demonstrated superior predictive performance, remarkable robustness in data-scarce regimes, and generated interpretable graphs aligned with domain logic.
Future work will explore scaling the joint estimation algorithm to massive dimensionalities via randomized or sketched precision estimators, as well as extending the PNN framework to temporal settings where the conditional dependency structure evolves over time.

\bibliographystyle{IEEEbib}
\bibliography{myIEEEabrv,bibliography} 

@article{sihag2022covariance,
  title={CoVariance neural networks},
  author={Sihag, S. and Mateos, G. and McMillan, C. T. and Ribeiro, A.},
  journal=NeurIPS,
  volume={35},
  pages={17003--17016},
  year={2022}
}

@article{deshp2016sparse,
  title={Sparse {PCA} via covariance thresholding},
  author={Deshpande, Y. and Montanari, A.},
  journal=JMLR,
  volume={17},
  number={141},
  pages={1--41},
  year={2016}
}

@article{gama2020stability,
  title={Stability properties of graph neural networks},
  author={Gama, F. and Bruna, J. and Ribeiro, A.},
  journal=IEEE_J_SP,
  volume={68},
  pages={5680--5695},
  year={2020},
  publisher={IEEE}
}

@ARTICLE{wang2023mhagnn,
  author={Wang, Y. and Wang, X. and Yang, H. and others},
  journal=IEEE_J_IM, 
  title={MhaGNN: A Novel Framework for Wearable Sensor-Based Human Activity Recognition Combining Multi-Head Attention and Graph Neural Networks}, 
  year={2023},
  volume={72},
  number={},
  pages={1-14},
  doi={10.1109/TIM.2023.3276004}}

@INPROCEEDINGS{liao2022har,
  author={Liao, T. and Zhao, J. and Liu, Y. and Ivanov, K. and Xiong, J. and Yan, Y.},
  booktitle={{IEEE} Int. Conf. Bioinform. Biomed.}, 
  title={Deep Transfer Learning with Graph Neural Network for Sensor-Based Human Activity Recognition}, 
  year={2022},
  volume={},
  number={},
  pages={2445-2452},
  doi={10.1109/BIBM55620.2022.9995660}}

@article{sihag2024explainable,
  title={Explainable brain age prediction using covariance neural networks},
  author={Sihag, S. and Mateos, G. and McMillan, C. T. and Ribeiro, A.},
  journal=NeurIPS,
  volume={36},
  year={2024}
}

@article{friedman2007lasso,
    author = {Friedman, J. and Hastie, T. and Tibshirani, R.},
    title = "{Sparse inverse covariance estimation with the graphical lasso}",
    journal = {Biostatistics},
    volume = {9},
    number = {3},
    pages = {432-441},
    year = {2007},
    month = {12},
    issn = {1465-4644},
    doi = {10.1093/biostatistics/kxm045},
    url = {https://doi.org/10.1093/biostatistics/kxm045},
    eprint = {https://academic.oup.com/biostatistics/article-pdf/9/3/432/57071116/biostatistics\_9\_3\_432.pdf},
}

@inproceedings{cavallo2024stvnn,
  title={Spatiotemporal covariance neural networks},
  author={Cavallo, A. and Sabbaqi, M. and Isufi, E.},
  booktitle={Joint Eur. Conf. Mach. Learn. Knowl. Discov. Databases},
  pages={18--34},
  year={2024},
  organization={Springer}
}

@book{Jolliffe2002pca,
  address = {New York},
  author = {Jolliffe, I. T.},
  publisher = {Springer Verlag},
  title = {Principal component analysis},
  year = 2002
}

@misc{misc_realdisp_activity_recognition_dataset_305,
  author       = {Banos, O. and Toth, M. and Amft, O.},
  title        = {{REALDISP Activity Recognition Dataset}},
  year         = {2014},
  howpublished = {UCI Machine Learning Repository},
  note         = {{DOI}: https://doi.org/10.24432/C5GP6D}
}

@misc{tello2023good,
      title={Too Good To Be True: performance overestimation in (re)current practices for Human Activity Recognition}, 
      author={Tello, A. and Degeler, V. and Lazovik, A.},
      year={2023},
      howpublished={arXiv preprint arXiv:2310.11950},
      eprint={2310.11950},
      archivePrefix={arXiv},
      primaryClass={cs.LG}
}

@inproceedings{banos2014mhealth,
author = {Banos, O. and Garcia, R. and Holgado-Terriza, J. and others},
year = {2014},
month = {12},
pages = {91--98},
title = {{mHealthDroid}: A Novel Framework for Agile Development of Mobile Health Applications},
volume = {8868},
isbn = {978-3-319-13104-7},
doi = {10.1007/978-3-319-13105-4_14}
}

@article{Gama_2020,
   title={Graphs, Convolutions, and Neural Networks: From Graph Filters to Graph Neural Networks},
   volume={37},
   ISSN={1558-0792},
   url={http://dx.doi.org/10.1109/MSP.2020.3016143},
   DOI={10.1109/msp.2020.3016143},
   number={6},
   journal={{IEEE} Signal Process. Mag.},
   publisher={Institute of Electrical and Electronics Engineers (IEEE)},
   author={Gama, F. and Isufi, E. and Leus, G. and Ribeiro, A.},
   year={2020},
   month=nov, pages={128–138} }

@article{Jolliffe2016PrincipalCA,
  title={Principal component analysis: a review and recent developments},
  author={Jolliffe, I. T. and Cadima, J.},
  journal={Philos. Trans. Roy. Soc. A, Math., Phys. Eng. Sci.},
  year={2016},
  volume={374},
}

@article{sihag2023transferablility,
  title={Transferablility of covariance neural networks and application to interpretable brain age prediction using anatomical features},
  author={Sihag, S. and Mateos, G. and McMillan, C. T. and Ribeiro, A.},
  journal={arXiv preprint arXiv:2305.01807},
  year={2023}
}

@article{cardoso2020algorithms,
  title={Algorithms for learning graphs in financial markets},
  author={Cardoso, J. V. and Ying, J. and Palomar, D. P.},
  journal={arXiv preprint arXiv:2012.15410},
  year={2020}
}

@inproceedings{wang2022network,
author = {Wang, Y. and Aste, T.},
title = {Network Filtering of Spatial-temporal {GNN} for Multivariate Time-series Prediction},
year = {2022},
isbn = {9781450393768},
publisher = {Association for Computing Machinery},
address = {New York, NY, USA},
url = {https://doi.org/10.1145/3533271.3561678},
doi = {10.1145/3533271.3561678},
booktitle = {{ACM} Int. Conf. AI Finance},
pages = {463--470},
numpages = {8},
location = {New York, NY, USA},
series = {ICAIF '22}
}

@book{vershynin2018high,
  title={High-Dimensional Probability: An Introduction with Applications in Data Science},
  author={Vershynin, R.},
  isbn={9781108415194},
  lccn={2018016910},
  series={Cambridge Series in Statistical and Probabilistic Mathematics},
  year={2018},
  publisher={Cambridge Univ. Press}
}

@ARTICLE{cavallo2024sparse,
  author={Cavallo, A. and Gao, Z. and Isufi, E.},
  journal={IEEE Trans. Signal Process.}, 
  title={Sparse Covariance Neural Networks}, 
  year={2026},
  volume={74},
  number={},
  pages={3077-3091},
  doi={10.1109/TSP.2026.3714283}}

@string{aaai = {AAAI Conf. on Artif. Intell.}}

@string{asilomar = {Asilomar Conf. Signals, Syst., and Computers}}

@string{icaif = {Intl. Conf. on AI in Finance}}

@string{mlsp = {Intl. Wrkshp. on Mach. Learn. for Signal Process. (MLSP)}}

@string{plos = {PLoS, Computat. Biology}}

@article{ravikumar2008estimationbounds,
      title={High-dimensional covariance estimation by minimizing $\ell_1$-penalized log-determinant divergence},
      author={Ravikumar, P. and Wainwright, M. J. and Raskutti, G. and Yu, B.},
      journal={Electron. J. Statist.},
      volume={5},
      pages={935--980},
      year={2011},
}

@article{navarro2024fair,
  title={Fair glasso: {E}stimating fair graphical models with unbiased statistical behavior},
  author={Navarro, M. and Rey, S. and Buciulea, A. and Marques, A. G. and Segarra, S.},
  journal=NeurIPS,
  volume={37},
  pages={139589--139620},
  year={2024}
}

@ARTICLE{giannakis2018topoid,
  author={Giannakis, G. B. and Shen, Y. and Karanikolas, G. V.},
  journal=IEEE_J_PROC, 
  title={Topology Identification and Learning over Graphs: Accounting for Nonlinearities and Dynamics}, 
  year={2018},
  volume={106},
  number={5},
  pages={787-807},
  doi={10.1109/JPROC.2018.2804318}}

@article{jack2008alzheimer,
  title={The {A}lzheimer's disease neuroimaging initiative ({ADNI}): {MRI} methods},
  author={Jack Jr, C. R. and Bernstein, M. A. and Fox, N. C. and Thompson, P. and Alexander, G. and Harvey, D. and Borowski, B. and Britson, P. J. and Whitwell, J. and Ward, C. and others},
  journal={J. Magn. Reson. Imaging},
  volume={27},
  number={4},
  pages={685--691},
  year={2008},
  publisher={Wiley Online Library}
}

@article{craddock2013neuro,
  title={The {N}euro {B}ureau preprocessing initiative: open sharing of preprocessed neuroimaging data and derivatives},
  author={Craddock, C. and Benhajali, Y. and Chu, C. and Chouinard, F. and Evans, A. and Jakab, A. and Khundrakpam, B. S. and Lewis, J. D. and Li, Q. and Milham, M. and others},
  journal={Front. Neuroinform.},
  volume={7},
  number={27},
  pages={5},
  year={2013}
}

@article{kingma2014adam,
  title={Adam: A method for stochastic optimization},
  author={Kingma, D. P. and Ba, J.},
  journal={arXiv preprint arXiv:1412.6980},
  year={2014}
}

@article{fischl2012freesurfer,
  title={FreeSurfer},
  author={Fischl, B.},
  journal={NeuroImage},
  volume={62},
  number={2},
  pages={774--781},
  year={2012},
  publisher={Elsevier}
}

@article{klein2017mindboggling,
  title={Mindboggling morphometry of human brains},
  author={Klein, A. and Ghosh, S. S. and Bao, F. S. and Giard, J. and H{\"a}me, Y. and Stavsky, E. and Lee, N. and Rossa, B. and Reuter, M. and Chaibub Neto, E. and others},
  journal={PLoS Comput. Biol.},
  volume={13},
  number={2},
  pages={e1005350},
  year={2017},
  publisher={Public Library of Science San Francisco, CA USA}
}

@book{lauritzen1996graphical,
    title={{Graphical Models}},
    author={Lauritzen, S. L.},
    volume={17},
    year={1996},
    publisher={Clarendon Press}
}

@ARTICLE{rey2023robust,
  author={Rey, S. and Tenorio, V. M. and Marques, A. G.},
  journal=IEEE_J_SP, 
  title={Robust Graph Filter Identification and Graph Denoising From Signal Observations}, 
  year={2023},
  volume={71},
  number={},
  pages={3651-3666},
  doi={10.1109/TSP.2023.3300632}}

@INPROCEEDINGS{tenorio2023robust,
  author={Tenorio, V. M. and Rey, S. and Marques, A. G.},
  booktitle={Asilomar Conf. Signals, Syst., Comput.}, 
  title={Robust Graph Neural Network Based on Graph Denoising}, 
  year={2023},
  volume={},
  number={},
  pages={578-582},
  doi={10.1109/IEEECONF59524.2023.10476925}}

@inproceedings{jin2020graph,
  title={Graph structure learning for robust graph neural networks},
  author={Jin, W. and Ma, Y. and Liu, X. and Tang, X. and Wang, S. and Tang, J.},
  booktitle={Intl. Conf. Knowl. Discovery Data Mining (ACM SIGKDD)},
  pages={66--74},
  year={2020}
}

@book{beck2017first,
  title={First-order methods in optimization},
  author={Beck, A.},
  year={2017},
  publisher={SIAM}
}

@inproceedings{natali2020topology,
  title={Topology-aware joint graph filter and edge weight identification for network processes},
  author={Natali, A. and Coutino, M. and Leus, G.},
  booktitle={{IEEE} Int. Workshop Mach. Learn. Signal Process. ({MLSP})},
  pages={1--6},
  year={2020},
  organization={IEEE}
}

@article{egilmez2018graph,
  title={Graph learning from filtered signals: Graph system and diffusion kernel identification},
  author={Egilmez, H. E. and Pavez, E. and Ortega, A.},
  journal={{IEEE} Trans. Signal, Inform. Process. Networks},
  volume={5},
  number={2},
  pages={360--374},
  year={2018},
  publisher={IEEE}
}

@inproceedings{cavallo2025pnns,
  title={Precision neural networks: Joint graph and relational learning},
  author={Cavallo, A. and Rey, S. and Marques, A. G. and Isufi, E.},
  booktitle={IEEE Int. Conf. Acoust., Speech and Signal Process.},
  pages={466--470},
  year={2026},
  organization={IEEE}
}

@article{zhang2022asgnn,
  title={{ASGNN}: Graph neural networks with adaptive structure},
  author={Zhang, Z. and Lu, S. and Huang, Z. and Zhao, Z.},
  journal={arXiv preprint arXiv:2210.01002},
  year={2022}
}

@inproceedings{wang2021graph,
  title={Graph structure estimation neural networks},
  author={Wang, R. and Mou, S. and Wang, X. and Xiao, W. and Ju, Q. and Shi, C. and Xie, X.},
  booktitle={Proc. Web Conf.},
  pages={342--353},
  year={2021}
}

@article{marek2011parkinson,
  title={The {P}arkinson progression marker initiative ({PPMI})},
  author={Marek, K. and Jennings, D. and Lasch, S. and Siderowf, A. and Tanner, C. and Simuni, T. and Coffey, C. and Kieburtz, K. and Flagg, E. and Chowdhury, S. and others},
  journal={Prog. Neurobiol.},
  volume={95},
  number={4},
  pages={629--635},
  year={2011},
  publisher={Elsevier}
}

@article{sihag2025disentangling,
  title={Disentangling Neurodegeneration with Brain Age Gap Prediction Models: A Graph Signal Processing Perspective},
  author={Sihag, S. and Mateos, G. and Ribeiro, A.},
  journal=IEEE_M_SP,
  volume={42},
  number={4},
  pages={58--77},
  year={2025},
  publisher={IEEE}
}

@article{yin2023anatomically,
  title={Anatomically interpretable deep learning of brain age captures domain-specific cognitive impairment},
  author={Yin, C. and Imms, P. and Cheng, M. and Amgalan, A. and Chowdhury, N. F. and Massett, R. J. and Chaudhari, N. N. and Chen, X. and Thompson, P. M. and Bogdan, P. and others},
  journal={Proc. Natl. Acad. Sci. USA},
  volume={120},
  number={2},
  pages={e2214634120},
  year={2023},
  publisher={National Academy of Sciences}
}

@article{bashyam2020mri,
  title={{MRI} signatures of brain age and disease over the lifespan based on a deep brain network and 14 468 individuals worldwide},
  author={Bashyam, V. M. and Erus, G. and Doshi, J. and Habes, M. and Nasrallah, I. M. and Truelove-Hill, M. and Srinivasan, D. and Mamourian, L. and Pomponio, R. and Fan, Y. and others},
  journal={Brain},
  volume={143},
  number={7},
  pages={2312--2324},
  year={2020},
  publisher={Oxford University Press}
}

@article{
roy2026covariance,
title={Covariance Density Neural Networks},
author={Roy, O. and Moshfeghi, Y. and Smith, K. M.},
journal={Trans. Mach. Learn. Res.},
issn={2835-8856},
year={2026},
url={https://openreview.net/forum?id=TwCkGi5XFB},
note={}
}

@inproceedings{cavallo2025covariancescattering,
  title={Covariance Scattering Transforms},
  author={Cavallo, A. and Raghuvanshi, A. and Chepuri, S. P. and Isufi, E.},
  booktitle={{AAAI} Conf. Artif. Intell.},
  volume={40},
  pages={19925--19933},
  year={2026}
}

@article{vowels2022d,
  title={D’ya like dags? a survey on structure learning and causal discovery},
  author={Vowels, M. J. and Camgoz, N. C. and Bowden, R.},
  journal={ACM Comput. Surv.},
  volume={55},
  number={4},
  pages={1--36},
  year={2022},
  publisher={ACM New York, NY}
}

@inproceedings{sahbi2021learning,
  title={Learning {L}aplacians in {C}hebyshev graph convolutional networks},
  author={Sahbi, H.},
  booktitle=ICCV,
  pages={2064--2075},
  year={2021}
}

@inproceedings{zhao2021heterogeneous,
  title={Heterogeneous graph structure learning for graph neural networks},
  author={Zhao, J. and Wang, X. and Shi, C. and Hu, B. and Song, G. and Ye, Y.},
  booktitle={{AAAI} Conf. Artif. Intell.},
  volume={35},
  pages={4697--4705},
  year={2021}
}

@article{chen2019deep,
  title={Deep iterative and adaptive learning for graph neural networks},
  author={Chen, Y. and Wu, L. and Zaki, M. J.},
  journal={arXiv preprint arXiv:1912.07832},
  year={2019}
}

@article{rothman2008sparse,
  title={Sparse permutation invariant covariance estimation},
  author={Rothman, A. J. and Bickel, P. J. and Levina, E. and Zhu, J.},
  journal={Electron. J. Statist.},
  volume={2},
  pages={494--515},
  year={2008},
  publisher={Institute of Mathematical Statistics}
}

@article{fatemi2021slaps,
  title={SLAPS: Self-supervision improves structure learning for graph neural networks},
  author={Fatemi, B. and El Asri, L. and Kazemi, S. M.},
  journal=NeurIPS,
  volume={34},
  pages={22667--22681},
  year={2021}
}

@inproceedings{franceschi2019learning,
  title={Learning discrete structures for graph neural networks},
  author={Franceschi, L. and Niepert, M. and Pontil, M. and He, X.},
  booktitle={Int. Conf. Mach. Learning},
  pages={1972--1982},
  year={2019},
  organization={PMLR}
}

@article{kittaneh1985lipschitz,
  title={On {L}ipschitz functions of normal operators},
  author={Kittaneh, F.},
  journal={Proc. Amer. Math. Soc.},
  volume={94},
  number={3},
  pages={416--418},
  year={1985}
}

@article{banerjee2008model,
  title={Model selection through sparse maximum likelihood estimation for multivariate {G}aussian or binary data},
  author={Banerjee, O. and El Ghaoui, L. and d'Aspremont, A.},
  journal=JMLR,
  volume={9},
  pages={485--516},
  year={2008}
}

@article{sihag2026learning,
  title={Learning with Covariance Matrices: Principal Component Analysis Meets Learning with Graphs},
  author={Sihag, S. and Cavallo, A. and Isufi, E. and Mateos, G. and Ribeiro, A.},
  journal={arXiv preprint arXiv:2609.10490},
  year={2026}
}

@article{zhang2026learning,
  title={Learning Graph Topology with Functional Priors via Bilevel Optimization},
  author={Zhang, C. and Liu, S. and Wai, H.-T. and So, A. M.-C.},
  journal={arXiv preprint arXiv:2606.13885},
  year={2026}
}

@article{saboksayr2021online,
  title={Online discriminative graph learning from multi-class smooth signals},
  author={Saboksayr, S. S. and Mateos, G. and Cetin, M.},
  journal={Signal Process.},
  volume={186},
  pages={108101},
  year={2021},
  publisher={Elsevier}
}

@article{batreddy2023robust,
  title={Robust graph learning for classification},
  author={Batreddy, S. and Siripuram, A. and Zhang, J.},
  journal={Signal Process.},
  volume={211},
  pages={109120},
  year={2023},
  publisher={Elsevier}
}

@inproceedings{raghuvanshi2025task,
  title={Task-driven Heterophilic Graph Structure Learning},
  author={Raghuvanshi, A. and Mateos, G. and Chepuri, S. P.},
  booktitle={Asilomar Conf. Signals, Syst., Comput.},
  pages={166--170},
  year={2025},
  organization={IEEE}
}

@STRING{IEEE_J_SP         = "{IEEE} Trans. Signal Process."}

@STRING{IEEE_J_PROC       = "Proc. {IEEE}"}

@STRING{IEEE_J_IM         = "{IEEE} Trans. Instrum. Meas."}

@STRING{IEEE_M_SP         = "{IEEE} Signal Process. Mag."}

@STRING{ASILOMAR	   = "Asilomar Conf. Signals, Systems and Comput."}

@STRING{ICCV		   = "{IEEE} Int. Conf. Comput. Vision"}

@STRING{NeurIPS		   = "Conf. Neural Inform. Process. Syst."}

@STRING{AAAI		   = "Assoc. Advancement Artificial Intell."}

@STRING{ACM		       = "Assoc. Comput. Mach."}

@STRING{PMLR           = "Proc. Mach. Learning Res."}

@STRING{NOV		   = "Nov."}

@STRING{JMLR               = "J. Mach. Learn. Res."}

\appendices

\section{Auxiliary Lemma}
\begin{lemma}\label{lem:lipschitz}
Let $\mtA,\mtB \in \reals^{N\times N}$ be symmetric matrices and let $h:\mathcal{I}\to\reals$ be Lipschitz with constant $P$ on an interval $\mathcal{I}$ containing the eigenvalues of both $\mtA$ and $\mtB$. Then
\begin{equation}\label{eq:lemma_lipschitz}
    \| h(\mtA) - h(\mtB) \|_F \leq P \| \mtA - \mtB \|_F .
\end{equation}
Consequently, for a precision filter $\mtH(\cdot)$ with frequency response $h(\cdot)$ and any $\vcx\in\reals^N$,
\begin{equation}\label{eq:lemma_filter}
    \|\mtH(\mtA)\vcx - \mtH(\mtB)\vcx\|_2 \le P\|\mtA-\mtB\|_F\|\vcx\|_2 .
\end{equation}
\end{lemma}

\begin{myproof}
This is a classical result~\cite{kittaneh1985lipschitz}; we include a short proof for completeness.
Let $\mtA = \mtU \mtLambda_A \mtU^\Tr$ and $\mtB = \mtQ \mtLambda_B \mtQ^\Tr$ be the eigendecompositions of $\mtA$ and $\mtB$, with eigenvalues $\{a_i\}_{i=1}^N$ and $\{b_j\}_{j=1}^N$, and denote by $\vcu_i$ and $\vcq_j$ the $i$-th column of $\mtU$ and the $j$-th column of $\mtQ$, respectively. Since $h(\mtA) = \mtU h(\mtLambda_A)\mtU^\Tr$ and $h(\mtB) = \mtQ h(\mtLambda_B)\mtQ^\Tr$, we have
\begin{align}
    [\mtU^\Tr (h(\mtA) - h(\mtB)) \mtQ]_{ij} &= (h(a_i) - h(b_j))\, \vcu_i^\Tr \vcq_j, \nonumber\\
    [\mtU^\Tr (\mtA - \mtB) \mtQ]_{ij} &= (a_i - b_j)\, \vcu_i^\Tr \vcq_j. \nonumber
\end{align}
By the Lipschitz property, $|[\mtU^\Tr (h(\mtA) - h(\mtB)) \mtQ]_{ij}| \le P\, |[\mtU^\Tr (\mtA - \mtB) \mtQ]_{ij}|$ for all $i,j$. Summing the squares over $i,j$ and using that the Frobenius norm is invariant to orthogonal transformations yields \eqref{eq:lemma_lipschitz}. Finally, \eqref{eq:lemma_filter} follows from $\|(\mtH(\mtA)-\mtH(\mtB))\vcx\|_2 \le \|\mtH(\mtA)-\mtH(\mtB)\|_2\|\vcx\|_2 \le \|\mtH(\mtA)-\mtH(\mtB)\|_F\|\vcx\|_2$.
\end{myproof}

\section{Proof of Proposition~\ref{th:stability_sample}}
\label{sec:sample_stab_proof}

Let $\Delta\mtC = \hat{\mtC} - \mtC_0$. Since $T \ge N$, $\hat\mtC$ is invertible with probability one, so that $\hat\mtTheta = \hat\mtC^{-1}$ is well defined.
By Lemma~\ref{lem:lipschitz} with $\mtA = \hat\mtTheta$ and $\mtB = \mtTheta_0$,
\begin{equation}\label{eq:filter_stability_bound}
    \| \mtH(\hat{\mtTheta})\vcx - \mtH(\mtTheta_0)\vcx \|_2 \leq P \| \hat{\mtTheta} - \mtTheta_0 \|_F \|\vcx \|_2 .
\end{equation}
We now derive a bound for $\| \hat{\mtTheta} - \mtTheta_0 \|_F$.
From the identity
$\hat{\mtTheta} - \mtTheta_0 = \hat{\mtC}^{-1}(\mtC_0 - \hat{\mtC})\mtC_0^{-1}$
and the inequality $\| \mtA \mtB \|_F \leq \| \mtA \|_2 \| \mtB \|_F$, we bound the Frobenius norm of the error:
\begin{equation}\label{eq:frob_bound_steps}
        \| \hat{\mtTheta} - \mtTheta_0 \|_F \leq \| \hat{\mtC}^{-1} \|_2 \| \Delta\mtC \|_F \| \mtC_0^{-1} \|_2.
    \end{equation}
By Assumption~\ref{as:data_distr}, $\| \mtC_0^{-1} \|_2 = \| \mtTheta_0 \|_2 = 1/\lambda_\textnormal{min}$.
To bound $\| \hat{\mtC}^{-1} \|_2$, we use Weyl's inequality, $\lambda_\textnormal{min}(\hat\mtC) \ge \lambda_\textnormal{min}(\mtC_0) - \|\Delta\mtC\|_2$, and the assumption $\|\Delta\mtC\|_2 \le \lambda_\textnormal{min}/2$, which gives
    \begin{equation}\label{eq:bound_hatcinv}
        \|\hat{\mtC}^{-1}\|_2 = \lambda_\textnormal{min}(\hat\mtC)^{-1} \leq (\lambda_\textnormal{min} - \|\Delta\mtC\|_2)^{-1} \le 2\lambda_\textnormal{min}^{-1} .
    \end{equation}
To bound the covariance error, we use the non-asymptotic bound for sub-Gaussian distributions~\cite[Theorem 4.7.1]{vershynin2018high}, which states that
$\| \Delta\mtC \|_2 = \mathcal{O}_P(\|\mtC_0\|_2\sqrt{N/T})$, where we used that for $T \ge N$ the term $N/T$ is dominated by $\sqrt{N/T}$.
Since $\| \mtA \|_F \leq \sqrt{N} \| \mtA \|_2$ for any $N \times N$ matrix, we have
    \begin{equation}\label{eq:samplecov}
        \| \Delta\mtC \|_F = \mathcal{O}_P \left( \| \mtC_0 \|_2 \frac{N}{\sqrt{T}} \right) = \mathcal{O}_P \left( \lambda_\textnormal{max} \frac{N}{\sqrt{T}} \right).
    \end{equation}
Substituting \eqref{eq:bound_hatcinv} and \eqref{eq:samplecov} in \eqref{eq:frob_bound_steps} yields
\begin{equation}\label{eq:sample_precision_error}
        \| \hat{\mtTheta} - \mtTheta_0 \|_F = \mathcal{O}_P \left( 2\lambda_\textnormal{max}\lambda_\textnormal{min}^{-2} N T^{-1/2} \right).
    \end{equation}
Since $\lambda_\textnormal{min}, \lambda_\textnormal{max}$ are bounded (cf.~Assumption~\ref{as:data_distr}),
$\mathcal{O}_P$ absorbs them. Replacing \eqref{eq:sample_precision_error} in \eqref{eq:filter_stability_bound} completes the proof.
\hfill $\blacksquare$

\section{Proof of Proposition~\ref{th:stability_gl}}
\label{sec:glasso_stab_proof}

By Lemma~\ref{lem:lipschitz} with $\mtA = \mtTheta_{\textnormal{GL}}$ and $\mtB = \mtTheta_0$,
    $\| \mtH(\mtTheta_{\textnormal{GL}})\vcx - \mtH(\mtTheta_0)\vcx \|_2 \leq P \| \mtTheta_{\textnormal{GL}} - \mtTheta_0 \|_F \|\vcx\|_2$ .
Under Assumption~\ref{as:data_distr} and with $\beta_\textnormal{GL} \asymp \sqrt{\log N/T}$, standard GL results~\cite[Theorem 1]{rothman2008sparse} (see also~\cite[eq. (43a)]{ravikumar2008estimationbounds}) yield
\begin{equation}\label{eq:gl_rate}
    \| \mtTheta_{\textnormal{GL}} - \mtTheta_0 \|_F = \mathcal{O}_P \left( \sqrt{((N+S)\log{N})/T} \right),
\end{equation}
with probability approaching 1 as $N, T \rightarrow \infty$.
Substituting \eqref{eq:gl_rate} into the previous inequality proves the claim.
\hfill $\blacksquare$

\section{Proof of Theorem~\ref{th:precision_convergence_ext}}
\label{sec:proof_joint}

To simplify the notation within the proof, we will denote $\mtTheta^{(i)}$ as $\mtTheta$ and $\mttTheta^{(i)}$ as $\mttTheta$. We define $\mtDelta = \mtTheta - \mtTheta_0$.
The derivations hold for one specific iteration $i$.
Let $\mathcal{L}_\mtTheta(\mtTheta)$ denote the loss in~\eqref{eq:step1} for a fixed $\tilde{\mtTheta}$:
\begin{align}
    \mathcal{L}_\mtTheta(\mtTheta) = &(1-\alpha) (-\log \det(\mtTheta) + \text{tr}(\mthC\mtTheta) + \beta_\text{GL} \|\mtTheta_{\bar{\mathcal{D}}}\|_1) \nonumber \\
    &+ \frac{\gamma}{2} \| \mtTheta - \mttTheta \|_F^2 ,\label{eq:ltheta_full}
\end{align}
and let $\mathcal{C} = \{\mtTheta : \mtTheta \succ 0,\ \|\mtTheta\|_2 \le M\}$ denote the convex feasible set of~\eqref{eq:step1}. Note that $\mtTheta_0 \in \mathcal{C}$, since $\|\mtTheta_0\|_2 = 1/\lambda_\textnormal{min} \le M$ by assumption.

Since $\mathcal{L}_\mtTheta$ is convex and $\mtTheta$ is its minimizer over $\mathcal{C}$, the first-order optimality condition~\cite{beck2017first} guarantees the existence of a subgradient $\mtG(\mtTheta) \in \partial \mathcal{L}_\mtTheta(\mtTheta)$ such that $\operatorname{tr}(\mtG(\mtTheta)(\mtTheta' - \mtTheta)) \geq 0$ for every $\mtTheta' \in \mathcal{C}$.
We can compute the following expression for $\mtG(\mtTheta)$
\begin{align}
    \mtG(\mtTheta) = (1-\alpha)( -\mtTheta^{-1} + \mthC + \beta_\text{GL} \mtZ ) + \gamma(\mtTheta - \mttTheta),
\end{align}
where $\mtZ$ is the subgradient of $\|\mtTheta_{\bar{\mathcal{D}}}\|_1$ with entries defined as
\begin{equation}
    z_{ij} \begin{cases}
        = \operatorname{sign}([\mtTheta]_{ij}) & \text{if } i \neq j \text{ and } [\mtTheta]_{ij} \neq 0 \\
        \in [-1, 1] & \text{if } i \neq j \text{ and } [\mtTheta]_{ij} = 0. \\
        = 0 & \text{if } i = j
    \end{cases}
\end{equation}

Taking $\mtTheta' = \mtTheta_0 \in \mathcal{C}$ in the optimality condition gives $\operatorname{tr}(\mtG(\mtTheta)(\mtTheta_0 - \mtTheta)) = -\operatorname{tr}(\mtG(\mtTheta)\mtDelta) \geq 0$, that is,
\begin{align}
    0 &\geq \operatorname{tr}(\mtG(\mtTheta)\mtDelta) = (1-\alpha)\operatorname{tr}((-\mtTheta^{-1} + \mthC)\mtDelta)  \nonumber \\
    & +(1-\alpha)\beta_\text{GL} \operatorname{tr}(\mtZ\mtDelta) + \gamma \operatorname{tr}((\mtTheta - \tilde{\mtTheta})\mtDelta) \nonumber
\end{align}
(with equality if the constraint $\|\mtTheta\|_2 \le M$ is inactive, in which case $\mtG(\mtTheta) = 0$).
By adding and subtracting $\mtC_0 = \mtTheta_0^{-1}$, the terms are rearranged as
\begin{align}
    0 &\geq (1-\alpha) \Big( \underbrace{\operatorname{tr}((\mthC - \mtC_0)\mtDelta)}_{\textnormal{Term 1}} + \underbrace{\operatorname{tr}((\mtC_0 - \mtTheta^{-1})\mtDelta)}_{\textnormal{Term 2}} \nonumber \\
    & + \underbrace{\beta_{\text{GL}} \operatorname{tr}(\mtZ\mtDelta)}_{\textnormal{Term 3}} \Big) + \underbrace{\gamma \operatorname{tr}((\mtTheta - \tilde{\mtTheta})\mtDelta)}_{\textnormal{Term 4}}. \label{eq:opt_cond_terms}
\end{align}
We proceed to lower bound Terms 1-4. Let $r_T = \sqrt{((N+S)\log N)/T}$.

\smallskip
\noindent \textbf{Term 1.}
For Term 1, we use the result from~\cite[(22)]{navarro2024fair} (see also~\cite[proof of Theorem 1]{rothman2008sparse}): there exist some constants $c_1,c_2$ such that, with high probability, the following holds.
\begin{align}
    \operatorname{tr}((\mthC - \mtC_0)\mtDelta) \geq& -c_1 \sqrt{\frac{\log N}{T}} \| \mtTheta_\mathcal{\bar{D}} - (\mtTheta_0)_\mathcal{\bar{D}} \|_1 \nonumber \\
    & -c_2 r_T \| \mtTheta_\mathcal{D} - (\mtTheta_0)_\mathcal{D} \|_F \label{eq:t1}
\end{align}
where $\mtTheta_\mathcal{D}$ and $\mtTheta_\mathcal{\bar{D}}$ are matrices containing the diagonal and off-diagonal values of $\mtTheta$, respectively.

\smallskip
\noindent \textbf{Term 2.}
Using $\mtC_0 = \mtTheta_0^{-1}$ and the identity $\mtTheta_0^{-1} - \mtTheta^{-1} = \mtTheta_0^{-1}(\mtTheta - \mtTheta_0)\mtTheta^{-1} = \mtTheta_0^{-1}\mtDelta\mtTheta^{-1}$, Term 2 can be written as
\begin{align}
    \operatorname{tr}((\mtC_0 - \mtTheta^{-1})\mtDelta) &= \operatorname{tr}(\mtTheta_0^{-1}\mtDelta\mtTheta^{-1}\mtDelta) \nonumber \\
    &= \| \mtTheta_0^{-1/2}\mtDelta\mtTheta^{-1/2} \|_F^2 , \label{eq:term2_identity}
\end{align}
where we used the cyclic property of the trace and the fact that $\mtTheta_0$ and $\mtTheta$ are positive definite, hence admit symmetric positive definite square roots.
For positive definite $\mtA, \mtB$ and any $\mtX$, we have $\|\mtA^{1/2}\mtX\mtB^{1/2}\|_F \geq \sqrt{\lambda_\textnormal{min}(\mtA)\lambda_\textnormal{min}(\mtB)}\,\|\mtX\|_F$. Since $\lambda_\textnormal{min}(\mtTheta_0^{-1}) = \lambda_\textnormal{min}(\mtC_0) = \lambda_\textnormal{min}$ and $\lambda_\textnormal{min}(\mtTheta^{-1}) = 1/\|\mtTheta\|_2 \geq 1/M$ because $\mtTheta \in \mathcal{C}$, we obtain for Term 2:
\begin{align}
    \operatorname{tr}((\mtC_0 - \mtTheta^{-1})\mtDelta) \geq \frac{\lambda_\textnormal{min}}{M}\| \mtDelta \|^2_F.\label{eq:t2}
\end{align}

\smallskip
\noindent \textbf{Term 3.}
For Term 3, by definition of the subgradient $\mtZ \in \partial \|\mtTheta_\mathcal{\bar{D}}\|_1$, we have $\operatorname{tr}(\mtZ\mtDelta) \geq \| \mtTheta_\mathcal{\bar{D}} \|_1 - \| (\mtTheta_0)_\mathcal{\bar{D}} \|_1$. We then use the result in \cite[(30)]{navarro2024fair}:
\begin{align}
    \beta_{\text{GL}}\operatorname{tr}(\mtZ\mtDelta) &\geq \beta_{\text{GL}}(\| \mtTheta_\mathcal{\bar{D}} \|_1 - \| (\mtTheta_0)_\mathcal{\bar{D}}  \|_1) \nonumber\\
    &\geq \beta_{\text{GL}} ( \| \mtDelta_{\mathcal{\bar{D}} \cap \mathcal{\bar{S}}} \|_1 - \| \mtDelta_{\mathcal{\bar{D}} \cap \mathcal{S}} \|_1 )  \label{eq:t3}
\end{align}
where $\mathcal{S}$ selects the matrix elements corresponding to the non-zero off-diagonal entries of $\mtTheta_0$ and $\mathcal{\bar{S}}$ the complementary entries.

\smallskip
\noindent \textbf{Term 4.}
We lower bound Term 4 by exploiting the Cauchy-Schwarz inequality for the Frobenius inner product:
\begin{align}
    \gamma \operatorname{tr}((\mtTheta - \tilde{\mtTheta})\mtDelta) \geq -\gamma \| \mtTheta - \tilde{\mtTheta} \|_F \| \mtDelta \|_F.\label{eq:t4}
\end{align}

\smallskip
\noindent \textbf{Final bound.}
We now combine \eqref{eq:t1}, \eqref{eq:t2}, \eqref{eq:t3}, \eqref{eq:t4} in \eqref{eq:opt_cond_terms}:
\begin{align}
    0 \geq& (1-\alpha)\Big(-c_1 \sqrt{\frac{\log N}{T}} \| \mtDelta_\mathcal{\bar{D}} \|_1 - c_2 r_T \| \mtDelta_\mathcal{D} \|_F \nonumber \\
     & +\frac{\lambda_\textnormal{min}}{M}\| \mtDelta \|^2_F + \beta_{\text{GL}} ( \| \mtDelta_{\mathcal{\bar{D}} \cap \mathcal{\bar{S}}} \|_1 - \| \mtDelta_{\mathcal{\bar{D}} \cap \mathcal{S}} \|_1 ) \Big) \nonumber \\
     & - \gamma \| \mtTheta - \tilde{\mtTheta} \|_F \| \mtDelta \|_F. \label{eq:combined_bound}
\end{align}
Now, let us set $\beta_\text{GL} = (c_1/\epsilon_1) \sqrt{\log N/T}$ for some $\epsilon_1 < 1$ (this is the constant implicit in the assumption $\beta_\textnormal{GL}\asymp\sqrt{\log N/T}$ of the Theorem).
Since $\| \mtDelta_{\bar{\mathcal{D}}} \|_1 = \| \mtDelta_{\mathcal{\bar{D}} \cap \mathcal{S}} \|_1 + \| \mtDelta_{\mathcal{\bar{D}} \cap \mathcal{\bar{S}}} \|_1$, the terms involving $\ell_1$ norms in \eqref{eq:combined_bound} can be grouped as
\begin{align}
    & c_1\sqrt{\frac{\log N}{T}}\left( \Big(\frac{1}{\epsilon_1}-1\Big)\|\mtDelta_{\mathcal{\bar{D}} \cap \mathcal{\bar{S}}}\|_1 - \Big(\frac{1}{\epsilon_1}+1\Big)\|\mtDelta_{\mathcal{\bar{D}} \cap \mathcal{S}}\|_1 \right) \nonumber\\
    &\geq -c_1\Big(1+\frac{1}{\epsilon_1}\Big)\sqrt{\frac{\log N}{T}}\,\|\mtDelta_{\mathcal{\bar{D}} \cap \mathcal{S}}\|_1  \label{eq:l1_step}\\
    &\geq -c_1\Big(1+\frac{1}{\epsilon_1}\Big)\sqrt{\frac{S\log N}{T}}\,\|\mtDelta\|_F \geq -c_1\Big(1+\frac{1}{\epsilon_1}\Big)\, r_T\, \|\mtDelta\|_F, \nonumber
\end{align}
where the first inequality uses $\epsilon_1 < 1$ to drop a non-negative term, the second uses that $\mtDelta_{\mathcal{\bar{D}} \cap \mathcal{S}}$ has at most $S$ non-zero entries, so that $\|\mtDelta_{\mathcal{\bar{D}} \cap \mathcal{S}}\|_1 \leq \sqrt{S}\|\mtDelta_{\mathcal{\bar{D}} \cap \mathcal{S}}\|_F \leq \sqrt{S}\|\mtDelta\|_F$, and the last one uses $\sqrt{S} \leq \sqrt{N+S}$.
Replacing \eqref{eq:l1_step} in \eqref{eq:combined_bound} and using $\| \mtDelta_\mathcal{D} \|_F \leq \| \mtDelta \|_F$, we obtain
\begin{align}
    0 \geq & \| \mtDelta \|_F \Bigg[ \frac{(1-\alpha)\lambda_\textnormal{min}}{M} \| \mtDelta \|_F  \nonumber \\
    & - (1-\alpha)\left(c_1\Big(1+\frac{1}{\epsilon_1}\Big) + c_2\right) r_T -\gamma \| \mtTheta - \tilde{\mtTheta} \|_F \Bigg]. \label{eq:s2}
\end{align}
If $\| \mtDelta \|_F = 0$ this holds trivially. Otherwise, \eqref{eq:s2} implies that the term in square brackets must be non-positive, which yields
\begin{align}
    \| \mtDelta \|_F \leq \frac{M}{\lambda_\textnormal{min}} \Bigg[ &\left(c_1\Big(1+\frac{1}{\epsilon_1}\Big) + c_2\right) r_T + \frac{\gamma}{1-\alpha} \| \mtTheta - \tilde{\mtTheta} \|_F \Bigg]. \nonumber
\end{align}
Absorbing the constants $M/\lambda_\textnormal{min}$, $c_1$, $c_2$ and $\epsilon_1$ (which do not depend on $N$, $T$, $\gamma$ or $\alpha$) into the $\mathcal{O}_P$ notation and using the assumption $\| \mtTheta - \tilde{\mtTheta} \|_F \leq \xi$, we have
\begin{align}\label{eq:interm_bound}
    \| \mtDelta \|_F = \mathcal{O}_P \left( r_T + \frac{\gamma}{1-\alpha}\,\xi \right),
\end{align}
which concludes the proof.
\hfill $\blacksquare$

\section{Proof of Theorem~\ref{th:stability}}
\label{sec:proof_pnn}

By Lemma~\ref{lem:lipschitz} with $\mtA = \tilde{\mtTheta}^{(i)}$ and $\mtB = \mtTheta_0$,
    $\| \mtH(\tilde{\mtTheta}^{(i)})\vcx - \mtH(\mtTheta_0)\vcx \|_2 \leq P \| \tilde{\mtTheta}^{(i)} - \mtTheta_0 \|_F \|\vcx\|_2 $.
Then, we apply the triangle inequality
\begin{align}\nonumber
    \| \tilde{\mtTheta}^{(i)} - \mtTheta_0 \|_F &\leq \| \tilde{\mtTheta}^{(i)} - \mtTheta^{(i)} \|_F + \| \mtTheta^{(i)} - \mtTheta_0 \|_F \\
    &\leq \xi + \| \mtTheta^{(i)} - \mtTheta_0 \|_F. \nonumber
\end{align}
Finally, we substitute $\| \mtTheta^{(i)} - \mtTheta_0 \|_F$ from Theorem~\ref{th:precision_convergence_ext}, which yields $\| \tilde{\mtTheta}^{(i)} - \mtTheta_0 \|_F = \mathcal{O}_P\big( \sqrt{(N+S)\log N/T} + (1 + \frac{\gamma}{1-\alpha})\xi \big)$ and proves the claim.
\hfill $\blacksquare$

\end{document}